\documentclass[letterpaper,journal]{IEEEtran}
\usepackage[T1]{fontenc}
\usepackage{newtxtext}
\usepackage{needspace}
\usepackage{amsmath,amsfonts}
\usepackage{algorithmic}
\usepackage{algorithm}
\usepackage{array}
\usepackage{booktabs}
\usepackage[caption=false,font=normalsize,labelfont=sf,textfont=sf]{subfig}
\usepackage{textcomp}
\usepackage{stfloats}
\usepackage{cuted}
\makeatletter
\newenvironment{wideablation}{\begin{strip}\def\@captype{table}}{\end{strip}}
\makeatother

\usepackage{url}
\usepackage{verbatim}
\usepackage{graphicx}
\usepackage{multirow}
\usepackage{cite}
\usepackage{placeins}
\AddToHook{env/equation/before}{%
  \setlength{\abovedisplayskip}{6pt plus 2pt minus 2pt}%
  \setlength{\belowdisplayskip}{6pt plus 2pt minus 2pt}%
  \setlength{\abovedisplayshortskip}{3pt plus 1pt minus 1pt}%
  \setlength{\belowdisplayshortskip}{3pt plus 1pt minus 1pt}%
}

\begin{document}

\title{IR275K: A Benchmark for Infrared Multi-Frame Super-Resolution Toward Efficient Remote Sensing}

\author{Jie Deng\textsuperscript{1,3}, Heyang Wang\textsuperscript{1,3},
        Changxin Wang\textsuperscript{1,3}, Junkai Shen\textsuperscript{1,3},
        Hongyi Chen\textsuperscript{1,3}, Zhiping He\textsuperscript{2,3,\textdagger},
        Hongxing Qi\textsuperscript{1,3,\textdagger}, Xudong Zhang\textsuperscript{1,3,*}, and Jianyu Wang\textsuperscript{1,3} 
\thanks{\textsuperscript{1}Hangzhou Institute for Advanced Study,
        Hangzhou 310024, China.
        \textsuperscript{2}Shanghai Institute of Technical Physics of the Chinese Academy of Sciences, Shanghai 200000, China.
        \textsuperscript{3}University of Chinese Academy of Sciences,
        Beijing 100000, China.}
\thanks{This work was supported by Intelligent Preprocessing Technology for Spaceborne Remote Sensing Spectral Data (B02006C021035), Hangzhou Institute for Advanced Study, UCAS, China.}
\thanks{*Corresponding author:
        Xudong Zhang (zhangxudong@ucas.ac.cn).
        \textsuperscript{\textdagger}Co-corresponding authors:
        Zhiping He (hzping@mail.sitp.ac.cn),
        Hongxing Qi (Qihongxing@ucas.ac.cn).}
}

\maketitle

\begin{abstract}
Infrared multi-frame super-resolution (MFSR) exploits complementary observations to recover spatial detail, but its evaluation remains fragmented across datasets and protocols. A common benchmark is needed to evaluate reconstruction under weak texture, varying thermal contrast, and noise in the source imagery. It should also characterize reconstruction performance across different inter-frame displacement ranges. We introduce IR275K, a curated benchmark containing 594 infrared video sequences and 275,196 frames. It provides sequence-level train/validation/test splits and a reproducible $\times 4$ evaluation protocol. Evaluation spans single-frame and multi-frame reconstruction using seven learned baselines, the CGMamba reference model, and bicubic interpolation. All learned methods are evaluated under both bicubic degradation and synthetic mixed sensor noise. EDVR leads the overall ranking under bicubic degradation, whereas MSDTGP leads under mixed sensor noise. DBSR leads the high-displacement UAV subset under both settings. Controlled ablations using CGMamba show that real neighboring frames improve PSNR over center-frame repetition by 1.04~dB under bicubic degradation and 0.95~dB under mixed sensor noise. Center-frame conditioning also improves reconstruction under both settings. The benchmark thus reveals how reconstruction rankings and the benefits of neighboring frames depend on evaluation conditions. The complete IR275K dataset is publicly available at: \url{https://github.com/InfraRecon7/IR275K}.
\end{abstract}


\section{Introduction}
\IEEEPARstart{I}{nfrared} remote sensing from satellite, airborne, and low-altitude platforms supports applications ranging from maritime surveillance to emergency response. Limited detector resolution constrains spatial detail, while power and downlink budgets restrict processing and data transmission~\cite{ref46}. Multi-frame super-resolution (MFSR) exploits complementary information in successive observations of the same scene to recover spatial detail without changing the imaging hardware. However, the reconstruction gain depends on scene content and correspondence between frames. Evaluating infrared MFSR therefore requires joint assessment of reconstruction quality, computational cost, and neighboring-frame gains across acquisition conditions.

Related work in spaceborne optical sensing also addresses measurement quality under noise and transmission constraints. Examples include waveform filtering~\cite{ref50}, photon-counting noise modeling~\cite{ref51}, and real-time denoising~\cite{ref52}. Temperature-compensated calibration of Chang'e-6 infrared spectroscopy further illustrates the importance of thermal operating conditions~\cite{ref33}.

Existing evaluation suites such as REDS~\cite{ref24}, Vimeo-90K~\cite{ref25}, and DIV2K~\cite{ref26} primarily address visible-light video or single-image restoration. They provide limited coverage of the thermal contrast, sensor noise, weak texture~\cite{ref47}, and platform-induced motion characteristic of infrared video. These factors affect temporal correspondence and the information available for reconstruction. Published infrared MFSR results remain distributed across private datasets and differing evaluation protocols, making comparisons difficult. A shared benchmark is therefore needed to evaluate reconstruction across these conditions and characterize when neighboring frames provide useful information.

We introduce IR275K, a curated benchmark for infrared multi-frame super-resolution. It contains 594 video sequences totaling 275,196 frames, with sequence-level train, validation, and test splits; no frame from the same sequence appears in more than one partition. IR275K harmonizes three public infrared sources into paired LR--HR data for a common $\times 4$ reconstruction task. Evaluation covers bicubic degradation and synthetic mixed sensor noise, with shared source and imaging-condition groups. Holding scene content and partitions fixed, we compare method rankings and neighboring-frame gains with and without added noise. The noisy setting requires joint noise suppression and spatial detail recovery. Source- and condition-specific analyses reveal performance differences hidden by aggregate scores.

To complement cross-method evaluation, we introduce CGMamba, a 7.43M-parameter state-space reference for controlled analysis of temporal information use. Its Center-Guided CrossMamba (CGCM) module conditions neighboring-frame processing on the center frame. Comparisons with parameter-matched controls separate the benefit of neighboring-frame information from the effects of center conditioning and feature aggregation. Grouped comparisons further characterize where neighboring-frame information improves infrared reconstruction.

Our contributions are:

\begin{enumerate}
\item{IR275K, a curated infrared MFSR benchmark containing 594 sequences and 275,196 frames, with fixed sequence-level splits and a reproducible $\times 4$ evaluation protocol.}
\item{Evaluation of seven learned baselines, CGMamba, and bicubic interpolation, revealing how reconstruction rankings vary across degradation settings, sources, and imaging conditions.}
\item{A controlled study of temporal information use with CGMamba, quantifying neighboring-frame benefits across acquisition conditions and center-conditioning gains across two aggregation heads.}
\end{enumerate}

\section{Related Work}

\subsection{Datasets and Benchmarks}

Dataset design determines which aspects of super-resolution can be evaluated. DIV2K~\cite{ref26} provides high-resolution RGB images for single-image SR, establishing a common setting for spatial detail recovery. Once neighboring observations are available, evaluation must also account for temporal correspondence and information aggregation. Vimeo-90K~\cite{ref25} provides short sequences for video processing, while REDS~\cite{ref24} combines video data with degradation settings for deblurring and SR. The transition from images to sequences therefore expands the evaluation from spatial reconstruction to the use of temporal information.

Thermal infrared evaluation additionally requires data that capture thermal appearance and sensor noise. Existing resources address these properties through different tasks: SatVideoIRSDT~\cite{li2026satvideodataset,ref31} and SIATD~\cite{ref32} provide moving-target detection sequences, whereas UAV-TSR++~\cite{ref30} supports any-scale thermal image SR. Cai et al.~\cite{Cai2024} provide a thermal video denoising dataset and a physics-inspired noise generator. Adapting these resources to multi-frame SR requires consistent temporal inputs, LR--HR supervision, and degradation settings. IR275K organizes imagery from infrared detection and UAV SR sources into a common $\times4$ multi-frame reconstruction task, with sequence-level partitions and source- and condition-specific evaluation.

\subsection{Infrared and Multi-Frame Super-Resolution}

Single-frame SR learns spatial priors to recover detail from one observation. Convolutional and remote-sensing designs exploit local structure, residual learning, and self-similarity~\cite{ref1,ref2,ref3,ref4,ref7,ref8,ref9,ref6}, while non-local and sparse attention~\cite{ref5,ref10} and Transformer-based restoration~\cite{ref11,ref12,ref13,ref53,ref14,ref15,ref16} expand the available context. Prior selection further adapts reconstruction to image content: PortraitSR~\cite{PortraitSR} uses progressive structure and texture guidance for faces, whereas infrared methods exploit phase and spectral information in GPSMamba~\cite{ref37}, wavelet modulation in IRSRMamba~\cite{ref38}, and thermal-spectrum and foundation-model priors in DifIISR~\cite{ref39}. These approaches recover detail within a single observation; neighboring frames introduce an additional source of evidence.

Temporal observations offer cues beyond single-frame priors. In infrared detection, DeepPro~\cite{li2026probing} models temporal saliency and correlation, while DHiF~\cite{Li2026DHiF} uses dynamic high-frequency convolution for target--clutter discrimination. Multi-frame SR uses neighboring observations to reconstruct the target frame, making correspondence and aggregation central design choices. Motion compensation and deformable alignment establish correspondence in VESPCN~\cite{ref27}, satellite video SR~\cite{ref41}, DCUNet~\cite{ref40}, and VSRM~\cite{ref42}. ICNet~\cite{ICNet} further couples alignment and reconstruction through iterative collaboration, allowing the two processes to interact during reconstruction.

The benefit of alignment depends on the reliability of correspondence, which can be difficult to establish in weakly textured infrared imagery. Feature-space fusion offers another route to combining observations, as demonstrated by deep reparametrization of MAP-based multi-frame restoration~\cite{ref43}. Correspondence is not the only source of uncertainty: the degradation process also affects reconstruction. DDL-BSR~\cite{DDLBSR} uses reconstruction difficulty to guide spatially varying degradation-kernel estimation and incorporates both cues into blind SR. These directions distinguish the problem of combining observations from that of modeling their degradation.

\subsection{State-Space Models for Super-Resolution}

Multi-frame reconstruction increases the number of spatial features to process, motivating efficient models of long-range context. Mamba~\cite{ref17} and Vision Mamba~\cite{ref22} use selective state-space processing for sequence and visual representation learning. SR studies extend this approach through cross-scale and dynamic state modeling~\cite{ref18,ref19}, multi-scale and spatial--frequency processing~\cite{ref20,ref21,ref34}, and compact or hybrid architectures~\cite{ref35,ref36}. The extension to multiple observations introduces a further choice: how one frame conditions the processing of another. QMambaBSR~\cite{ref44} explores inter-frame querying for burst SR, while MamEVSR~\cite{ref45} uses interleaved and cross-modality Mamba blocks for event-based VSR. CGMamba uses the center frame to condition cross-frame feature processing, providing a reference for controlled study of neighboring-frame information and center conditioning.

\section{The IR275K Benchmark}

\subsection{Data Sources and Acquisition Diversity}

IR275K draws from three public infrared sources: SatVideoIRSDT~\cite{li2026satvideodataset,ref31}, the UAV-TSR++ dataset introduced with AnyTSR++~\cite{ref30}, and SIATD~\cite{ref32}. SatVideoIRSDT and SIATD were developed for infrared moving-target detection, whereas UAV-TSR++ targets any-scale UAV thermal image super-resolution. IR275K brings imagery from these resources into a common $\times 4$ multi-frame reconstruction task, with sequence-level partitions, paired LR--HR data, and consistent evaluation metrics. The benchmark combines bicubic and synthetic mixed-noise settings with source-wise and imaging-condition analyses. This unified framework supports reproducible comparisons across sources and degradation settings. Grouped evaluation further reveals performance differences and reconstruction challenges obscured by aggregate scores, informing the design and assessment of infrared MFSR methods.

Before quality screening, the available HR pools contain 165,882, 24,000, and 148,963 images or spatial crops, respectively. Detailed counts and crop accounting are provided in Supplementary Section~S2. The final frame contribution of each source depends on its available data volume and the outcome of quality screening.

\begin{figure*}[!t]
\centering
\includegraphics[width=\textwidth]{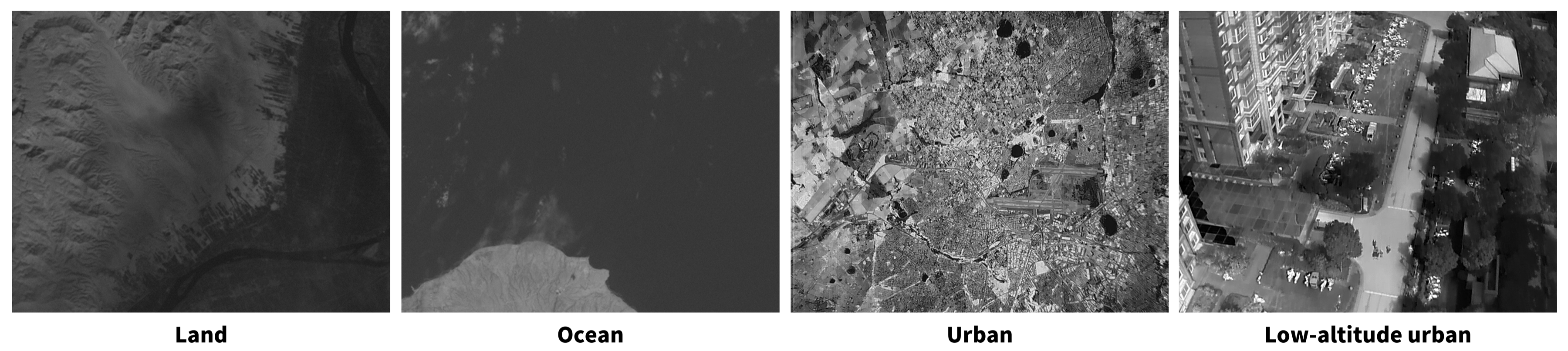}
\caption{Representative scenes in IR275K, including land remote sensing, ocean remote sensing, urban remote sensing, and low-altitude sensing. These examples illustrate the diversity of thermal appearance, target structure, and imaging context covered by the benchmark.}
\label{fig:ir275k_representative_scenes}
\end{figure*}

IR275K covers land, ocean, urban, and low-altitude scenes (Fig.~\ref{fig:ir275k_representative_scenes}). Scene labels are assigned at the sequence level using representative frames. The low-altitude sequences originate exclusively from UAV-TSR++.

\subsection{Curation Criteria}

Source frames were organized into sequences according to scene continuity and temporal correspondence. Sequence-level quality control excluded sky-dominated content, as well as sequences with insufficient structural detail or severe blur that made them unsuitable as HR references. Additional exclusions covered static sequences lacking useful inter-frame variation and sequences with scene cuts or discontinuous motion that disrupted temporal correspondence. Adjacent frame pairs with PSNR $\geq 50$~dB were flagged as duplicate or near-duplicate candidates.

The source-specific retained fractions are 86.34\% for SatVideoIRSDT, 99.56\% for UAV-TSR++, and 72.11\% for SIATD. Detailed source counts and retention calculations, including the accounting for spatial crops, are provided in Supplementary Table~S1 and Section~S2.

\subsection{Benchmark Statistics and Sequence-Level Partitioning}

\begin{figure*}[!t]
\centering
\includegraphics[width=\textwidth]{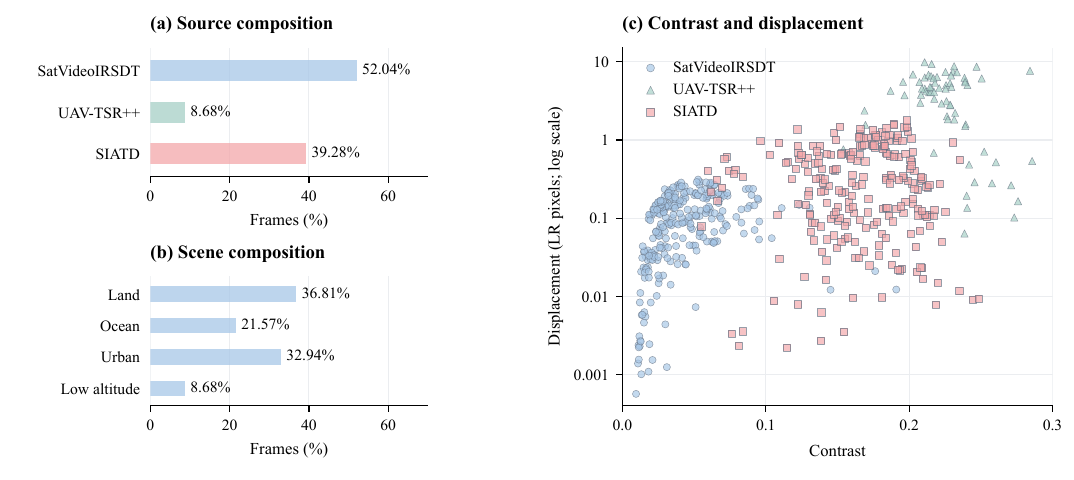}
\caption{Composition and imaging-condition coverage of IR275K. (a) Source and (b) scene contributions as percentages of all 275,196 frames. (c) Sequence-level contrast versus estimated inter-frame displacement for all 594 sequences. Colors and marker shapes in (c) identify data sources. Displacement is measured in native LR pixels on a logarithmic axis. Descriptors use the original data before synthetic noise addition.}
\label{fig:ir275k_dataset_statistics}
\end{figure*}

IR275K contains 594 sequences totaling 275,196 frames, as summarized in Table~\ref{tab:ir275k_stats} and Fig.~\ref{fig:ir275k_dataset_statistics}. SatVideoIRSDT contributes 283 sequences and 143,217 frames, accounting for 52.04\% of all frames; UAV-TSR++ contributes 67 sequences and 23,895 frames, accounting for 8.68\%; and SIATD contributes 244 sequences and 108,084 frames, accounting for 39.28\%. Sequence lengths range from 16 to 5,625 frames. The high-resolution reference frames are 640 $\times$ 512 pixels, and the corresponding 160 $\times$ 128 low-resolution inputs are generated by bicubic downsampling at $\times 4$ scale.

The benchmark was partitioned by randomly shuffling complete sequences and assigning them to the training, validation, and test sets. This sequence-level procedure prevents frames from the same original sequence from appearing in different splits. Training receives 261 sequences with 192,776 frames, validation receives 111 sequences with 27,576 frames, and test receives 222 sequences with 54,844 frames. The training split contains longer sequences on average, with a mean of 738.6 frames and a median of 548. Validation and test sequences average 248.4 and 247.0 frames, respectively.

\begin{table*}[!t]
\caption{Partition Sizes, Sequence Lengths, and Source Composition of IR275K. HR frames are stored at 640 $\times$ 512 pixels, the corresponding LR frames are 160 $\times$ 128 pixels, and all LR-HR pairs follow a uniform $\times 4$ scale.\label{tab:ir275k_stats}}
\centering
\begingroup
\small
\renewcommand{\arraystretch}{1.0}
\setlength{\tabcolsep}{4pt}
\begin{tabular*}{\linewidth}{@{\extracolsep{\fill}}cccccccc}
\toprule[1.2pt]
\multirow{2}{*}{\textbf{Split}} & \multirow{2}{*}{\textbf{Sequences}} & \multirow{2}{*}{\textbf{Frames}} & \multirow{2}{*}{\shortstack{\textbf{Mean}\\\textbf{length}}} & \multirow{2}{*}{\shortstack{\textbf{Median}\\\textbf{length}}} & \multicolumn{3}{c}{\textbf{Source contribution (\% of frames)}}\\
\cmidrule(lr){6-8}
 & & & & & \textbf{SatVideoIRSDT} & \textbf{UAV-TSR++} & \textbf{SIATD}\\
\midrule[0.6pt]
Train & 261 & 192,776 & 738.6 & 548 & 55.10 & 9.60 & 35.30\\
Validation & 111 & 27,576 & 248.4 & 300 & 49.45 & 6.94 & 43.61\\
Test & 222 & 54,844 & 247.0 & 300 & 42.60 & 6.33 & 51.06\\
\addlinespace[2pt]
Total & 594 & 275,196 & 463.3 & 400 & 52.04 & 8.68 & 39.28\\
\bottomrule[1.2pt]
\end{tabular*}
\par\smallskip
\begin{minipage}{\linewidth}
\footnotesize
Sequence lengths are measured in frames. Each source percentage is the number of frames from that source divided by the total number of frames in the corresponding row. Percentages are rounded to two decimal places. Minimum and maximum sequence lengths are reported in Supplementary Table~S3.
\end{minipage}
\endgroup
\end{table*}

\subsection{Split Characterization}
\label{sec:split_characterization}

After partitioning, we characterized the fixed training, validation, and test sets in terms of source and scene composition, thermal contrast, temporal variability, and inter-frame displacement. These complementary statistics describe acquisition composition, signal strength, measurement variability, and correspondence difficulty in the resulting benchmark partitions.

\emph{Source and scene composition.}
All three sources and all four scene categories are represented in the training, validation, and test sets. SatVideoIRSDT contributes 55.10\%, 49.45\%, and 42.60\% of the frames in the three partitions, respectively. The corresponding proportions are 9.60\%, 6.94\%, and 6.33\% for UAV-TSR++, and 35.30\%, 43.61\%, and 51.06\% for SIATD (Table~\ref{tab:ir275k_stats}). Detailed sequence and frame counts by source and scene category are provided in Supplementary Table~S2. Fig.~\ref{fig:ir275k_dataset_statistics} shows the overall source and scene composition alongside sequence-level contrast--displacement coverage by source.

\begin{figure*}[!t]
\centering
\includegraphics[width=\textwidth]{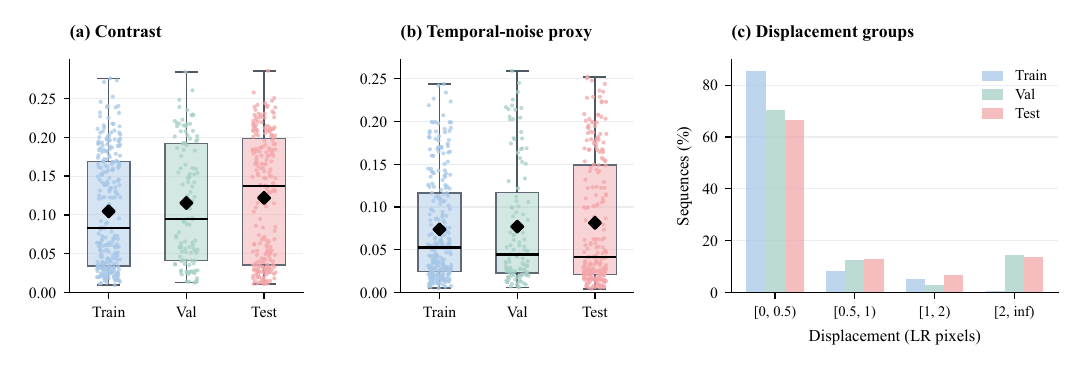}
\caption{Imaging-condition distributions across the IR275K partitions. (a) Contrast. (b) Temporal-noise proxy. Points represent sequences; boxes show interquartile ranges, horizontal lines medians, whiskers full ranges, and black diamonds means. Intensities are normalized to $[0,1]$. (c) Sequence percentages in native-LR-pixel displacement groups. Colors distinguish the 261 training, 111 validation, and 222 test sequences. All descriptors use the original data before synthetic noise addition.}
\label{fig:ir275k_split_characterization}
\label{fig:ir275k_contrast_noise}
\label{fig:ir275k_motion}
\end{figure*}

\emph{Thermal contrast.}
Contrast is measured as the sequence-average spatial standard deviation of full HR frames, with intensities normalized to $[0,1]$. Across IR275K, contrast ranges from 0.010 to 0.286. The training, validation, and test splits have sequence-mean values of 0.105, 0.115, and 0.122, respectively. The training interquartile range is [0.034, 0.169]. Fig.~\ref{fig:ir275k_split_characterization}(a) shows these distributions, using the same descriptor as the condition-specific evaluation in Section~\ref{sec:condition_performance}.

\emph{Temporal-noise proxy.}
The temporal-noise proxy is the spatial mean of per-pixel temporal standard deviations over all LR frames in a fixed central $64\times64$ region. Intensities are normalized to $[0,1]$, and population standard deviations are used. This proxy includes variability from noise, scene motion, and brightness changes. Across IR275K, it ranges from 0.004 to 0.259. The training, validation, and test splits have sequence-mean values of 0.074, 0.077, and 0.081, respectively. Their upper quartiles are 0.116, 0.117, and 0.149 (Fig.~\ref{fig:ir275k_split_characterization}(b)).

\emph{Inter-frame displacement.}
To define motion groups for condition-specific evaluation, we estimate inter-frame displacement from the original $160\times128$ bicubic LR frames, without additional resizing or added noise. Farneback optical flow is computed for adjacent frames within each sequence. Forward--backward consistency and texture checks identify reliable pixels; their median flow magnitude is averaged across valid frame pairs to obtain one descriptor per sequence. Split statistics give equal weight to each sequence. The descriptor and reliability criteria are detailed in Supplementary Section~S3.

For the fixed training, validation, and test partitions, the mean displacements are 0.265, 0.907, and 1.036~LR pixels, respectively. The training median and 95th percentile are 0.131 and 1.084~LR pixels. Validation and test maxima reach 7.67 and 9.85~LR pixels. Fig.~\ref{fig:ir275k_split_characterization}(c) reports the proportions of sequences in the same displacement intervals used for performance analysis. Group labels are retained for both degradation settings.

Source and scene composition, contrast, temporal variability, and inter-frame displacement characterize the conditions represented by IR275K. Section~\ref{sec:condition_performance} relates these descriptors to reconstruction performance through group-wise evaluation.

\subsection{Protocol, Reproducibility, and Scope}

IR275K defines a single evaluation protocol for all methods. The task is $\times 4$ upscaling of the center frame from consecutive low-resolution infrared inputs. Single-image methods use one input frame, while multi-frame methods use three.

Alongside bicubic degradation, IR275K includes a mixed-sensor-noise setting adapted from the physics-inspired thermal infrared noise generator of Cai et al.~\cite{Cai2024}. The mixture comprises pixelwise Gaussian noise, time-varying row noise, temporally shared row and column noise, and a spatial bias field. Their strengths are fixed at $(10,10,5,5,10)$ for training, validation, and testing. The first four values are Gaussian standard deviations; the fifth controls the bias-field amplitude. All values are in 8-bit-equivalent units and divided by 255 for normalized LR images.

During training, random noise is added online to sampled LR patches and resampled at every optimizer update. Row and column patterns are shared across input frames, while pixelwise Gaussian and time-varying row noise are sampled independently for distinct frames. Repeated boundary frames share the same noise realization. The bias field is defined in full-frame LR coordinates and undergoes the same cropping and horizontal flipping as the input images.

Validation and test inputs are generated once at full LR resolution and kept fixed across methods. Row and column patterns remain constant within each sequence, while the frame-varying components are sampled independently for each physical frame. In both training and evaluation, corrupted intensities are clipped to $[0,1]$ and quantized to 8-bit precision. HR targets and data partitions remain unchanged.

Both degradation settings are synthetic and do not model detector-specific blur or compression artifacts. The mixed-noise extension is not calibrated to a specific sensor. Evaluation is limited to PSNR, SSIM, and computational cost.

The imaging characteristics of IR275K motivate a reference architecture that aggregates neighboring-frame information in feature space. CGMamba combines per-frame spatial processing with center-conditioned cross-frame interaction and center-frame reconstruction. Controlled ablations then assess how much each source of information and conditioning contributes under the shared benchmark protocol.

\section{SSM Reference Model: CGMamba}
\subsection{Overall Architecture}

\begin{figure*}[!t]
  \centering
  \includegraphics[width=0.85\textwidth]{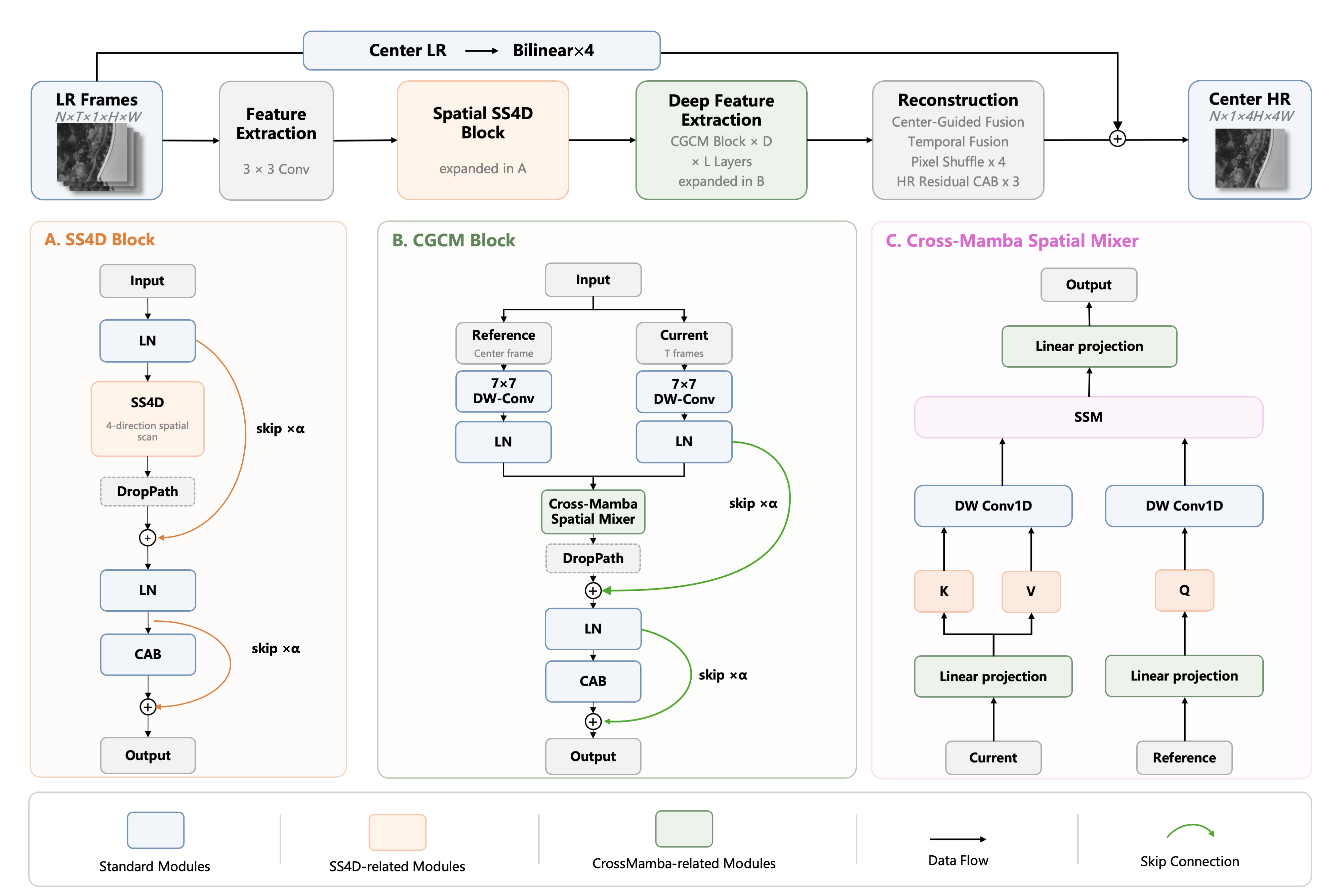}
  \caption{Overall architecture of CGMamba. (A) SS4D block. (B) CGCM block. (C) Cross-Mamba spatial mixer.}
  \label{fig:cgmamba_overall}
\end{figure*}

As shown in Fig.~\ref{fig:cgmamba_overall}, CGMamba reconstructs the high-resolution center frame from three low-resolution infrared frames. A $3\times3$ convolution and an SS4D module first extract spatial features independently from each frame. Center-Guided CrossMamba Groups then process these features using the center frame as the reference for neighboring-frame interaction. A terminal aggregation head produces a center-frame representation, followed by pixel-shuffle upsampling and residual channel-attention refinement at HR resolution.

A bilinearly upsampled center LR image is added to the predicted HR residual. CGMamba contains 7.43M parameters.

\subsection{Spatial Feature Processing}

The SS4D module scans each frame in four spatial directions and combines the directional responses before projection back to the feature space. Local convolutions and channel attention complement this spatial context modeling. Parameters are shared across frames, while the SS4D operations themselves process each frame independently. Cross-frame conditioning is introduced by the subsequent CGCM groups and terminal feature aggregation.

Within each CGCM group, the center-frame feature at the group entrance provides the reference for all blocks in that group. The reference is refreshed at the entrance to the next group. This group-level reference preserves a common center-frame conditioning signal while the current-frame features are progressively refined.

\subsection{Center-Guided CrossMamba}

In infrared MFSR, explicit alignment via optical flow or deformable convolution may become unreliable under weak texture and thermal homogeneity. We therefore adopt implicit fusion through reference-guided cross-frame gating: the center frame serves as a structural reference that guides feature extraction from each neighboring frame, without estimating an explicit motion field.

\subsubsection{Reference-Guided Cross-Frame Gating}

Fig.~\ref{fig:cgmamba_overall}(B) shows the CGCM block.

Using structured state-space duality (SSD)~\cite{ref28}, we express the output at spatial scan position $t$ as:
\begin{equation}
\label{eq:ssd_duality}
y_t = \sum_{s \leq t} C_t^{\top} \left( \prod_{k=s+1}^{t} A_k \right) B_s \, x_s.
\end{equation}
Here, $s$ and $t$ index spatial tokens within a frame, and $x_s$ and $y_t$ denote input and output features. The matrices $A_k$, $B_s$, and $C_t$ parameterize discrete state transition, input, and readout, respectively. The factor $C_t^{\top}(\prod A_k)B_s$ is structurally analogous to $Q_t^{\top}K_s$ in attention, motivating separate query and key pathways.

Let $F_i$, $i\in\{1,2,3\}$, denote the input-frame feature maps, and let $F_{\text{center}}$ denote the center-frame reference. Each current branch generates keys $K$, values $V$, and step sizes $\triangle$ from $F_i$, while the reference branch generates queries $Q$ from $F_{\text{center}}$:

\begin{equation}
\label{cgcm_kv}
K, V, \triangle = \text{Linear}_{\text{current}} \left( F_i \right)
\end{equation}

\begin{equation}
\label{cgcm_q}
Q = \text{Linear}_{\text{reference}} \left( F_{\text{center}} \right)
\end{equation}

The hidden state update and output are computed as: 

\begin{equation}
\label{eq:cgcm_scan}
\left\{
\begin{aligned}
h_t &= \exp\left(\triangle_t A\right) h_{t-1} + \triangle_t K_t V_t \\
y_t &= Q_t h_t
\end{aligned}
\right.
\end{equation}

Here, $h_t$ is the latent state at spatial position $t$, and $A$ governs state decay. The input-dependent step size $\triangle_t$ controls state discretization.

The reference branch conditions the state readout on center-frame features, while each current branch supplies its own keys, values, and state dynamics. This construction differs from processing every frame with a self-conditioned SSM and combining the resulting features only at the output. Inference uses the chunked recurrent SSM form, whose computational cost scales linearly with sequence length for fixed state and chunk sizes. Section~\ref{sec:controlled_ablation} evaluates the center-conditioned and self-conditioned alternatives with the remaining backbone components retained.

\subsection{Training Objective and Design Rationale}

We train CGMamba using the Charbonnier loss~\cite{ref29}:

\begin{equation}
\label{eq:rec_loss}
\mathcal{L}_{rec} = \sqrt{(I_{SR} - I_{GT})^2 + \epsilon^2}
\end{equation}

with $\epsilon = 10^{-3}$. The Charbonnier loss interpolates between $\ell_2$ and $\ell_1$, behaving quadratically near zero for stable convergence and linearly for large residuals to suppress outliers.

\section{Experiments}
\subsection{Experimental Setup}
\label{sec:experimental_setup}
CGMamba uses depths $(3,3,3,3,3)$ and feature dimension 128. The benchmark training protocol follows the CGMamba implementation, using $32\times32$ LR patches, an effective batch size of 128, and AdamW with a peak learning rate of $10^{-4}$. Training is performed separately under bicubic and mixed-noise degradation, with a budget of 100,000 updates. Evaluation uses the 100K EMA weights.

\begin{table*}[!t]
\caption{Benchmark Results on IR275K Under Bicubic and Mixed-Noise Degradation}
\label{tab:sota_comparison}
\label{tab:condition_performance}
\centering
\begingroup
\footnotesize
\renewcommand{\arraystretch}{0.95}
\setlength{\tabcolsep}{3pt}
\textbf{(a) Overall performance and computational cost}\par\smallskip
\begin{tabular*}{\textwidth}{@{\extracolsep{\fill}}ccccccc@{}}
\toprule[1.2pt]
\textbf{Method} & \textbf{Frames} & \shortstack{\textbf{Params (M)}$\downarrow$} & \shortstack{\textbf{FLOPs (G)}$\downarrow$} & \shortstack{\textbf{Time (ms)}$\downarrow$} & \shortstack{\textbf{Bicubic}\\\textbf{PSNR / SSIM}$\uparrow$} & \shortstack{\textbf{Bicubic + noise}\\\textbf{PSNR / SSIM}$\uparrow$}\\
\midrule[0.6pt]
Bicubic & 1 & -- & -- & 0.072 & 31.56\,/\,0.8182 & 23.28\,/\,0.4965\\
GPSMamba & 1 & 36.94 & 346.65 & 582.10 & 32.86\,/\,0.8483 & 28.78\,/\,0.7978\\
IRSRMamba & 1 & 26.61 & 245.56 & 179.25 & 32.67\,/\,0.8447 & 28.37\,/\,0.7793\\
CGMamba & 3 & 7.43 & 232.43 & 97.42 & 33.83\,/\,0.8763 & 29.91\,/\,0.8169\\
EDVR~\cite{EDVR} & 3 & 20.57 & 228.88 & 26.96 & \textbf{34.14}\,/\,\textbf{0.8854} & 29.58\,/\,0.8184\\
BasicVSR++~\cite{BasicVSRPP} & 3 & 7.32 & 163.89 & 36.36 & 34.06\,/\,0.8841 & 30.06\,/\,0.8208\\
DBSR~\cite{DBSR} & 3 & 12.91 & 86.46 & 25.75 & 33.83\,/\,0.8772 & 29.99\,/\,0.8184\\
MSDTGP~\cite{ref41} & 3 & 14.16 & 502.83 & 120.00 & 33.86\,/\,0.8778 & \textbf{30.13}\,/\,\textbf{0.8210}\\
MADNet~\cite{MADNet} & 3 & 9.17 & 240.07 & 57.16 & 33.74\,/\,0.8763 & 29.38\,/\,0.8062\\
\bottomrule[1.2pt]
\end{tabular*}
\par\medskip
\textbf{(b) Performance across sources and imaging conditions}\par\smallskip
\renewcommand{\arraystretch}{1.0}
\setlength{\tabcolsep}{2pt}
\begingroup
\scriptsize
\setlength{\tabcolsep}{1.5pt}
\begin{tabular*}{\textwidth}{@{\extracolsep{\fill}}ccccccccccccc@{}}
\toprule[1.2pt]
\textbf{Factor} & \textbf{Group} & \shortstack{\textbf{Sequences}\\\textbf{/ Frames}} & \textbf{Setting} & \shortstack{\textbf{Bicubic}} & \shortstack{\textbf{GPS}\\\textbf{Mamba}} & \shortstack{\textbf{IRSR}\\\textbf{Mamba}} & \shortstack{\textbf{CG}\\\textbf{Mamba}} & \shortstack{\textbf{EDVR}} & \shortstack{\textbf{Basic}\\\textbf{VSR++}} & \shortstack{\textbf{DBSR}} & \shortstack{\textbf{MSDTGP}} & \shortstack{\textbf{MADNet}}\\
\midrule[0.6pt]
\multirow{6}{*}{Source}  &  \multirow{2}{*}{\shortstack{SatVideo\\IRSDT}}  &  \multirow{2}{*}{97 / 23,366}  &  Bicubic  & \shortstack{38.55\\0.9082} & \shortstack{39.17\\0.9164} & \shortstack{38.98\\0.9143} & \shortstack{40.14\\0.9285} & \shortstack{\textbf{40.18}\\\textbf{0.9289}} & \shortstack{40.12\\0.9281} & \shortstack{40.02\\0.9270} & \shortstack{40.12\\0.9282} & \shortstack{39.95\\0.9259}\\
  &    &    &  Bicubic + noise  & \shortstack{24.76\\0.5061} & \shortstack{32.40\\0.8745} & \shortstack{31.82\\0.8481} & \shortstack{33.98\\0.8815} & \shortstack{32.90\\0.8693} & \shortstack{34.20\\0.8845} & \shortstack{34.03\\0.8819} & \shortstack{\textbf{34.31}\\\textbf{0.8870}} & \shortstack{33.23\\0.8735}\\
\addlinespace[2pt]
  &  \multirow{2}{*}{UAV-TSR++}  &  \multirow{2}{*}{36 / 3,472}  &  Bicubic  & \shortstack{24.05\\0.7311} & \shortstack{26.07\\0.8013} & \shortstack{25.79\\0.7941} & \shortstack{26.08\\0.8010} & \shortstack{26.60\\0.8162} & \shortstack{\textbf{26.94}\\0.8243} & \shortstack{26.93\\\textbf{0.8277}} & \shortstack{26.13\\0.8026} & \shortstack{26.50\\0.8158}\\
  &    &    &  Bicubic + noise  & \shortstack{21.40\\0.4810} & \shortstack{24.64\\0.7110} & \shortstack{24.38\\0.7118} & \shortstack{24.68\\0.7198} & \shortstack{25.07\\0.7380} & \shortstack{24.83\\0.7280} & \shortstack{\textbf{25.44}\\\textbf{0.7523}} & \shortstack{24.82\\0.7300} & \shortstack{24.84\\0.7282}\\
\addlinespace[2pt]
  &  \multirow{2}{*}{SIATD}  &  \multirow{2}{*}{89 / 28,006}  &  Bicubic  & \shortstack{26.66\\0.7539} & \shortstack{28.44\\0.7972} & \shortstack{28.26\\0.7930} & \shortstack{29.53\\0.8421} & \shortstack{\textbf{30.03}\\\textbf{0.8577}} & \shortstack{29.89\\0.8548} & \shortstack{29.51\\0.8418} & \shortstack{29.60\\0.8451} & \shortstack{29.46\\0.8425}\\
  &    &    &  Bicubic + noise  & \shortstack{22.28\\0.4904} & \shortstack{26.28\\0.7446} & \shortstack{25.98\\0.7303} & \shortstack{27.17\\0.7751} & \shortstack{\textbf{27.38}\\\textbf{0.7858}} & \shortstack{27.25\\0.7792} & \shortstack{27.19\\0.7737} & \shortstack{27.30\\0.7771} & \shortstack{26.72\\0.7597}\\
\addlinespace[5pt]
\multirow{6}{*}{Contrast}  &  \multirow{2}{*}{Low}  &  \multirow{2}{*}{74 / 16,669}  &  Bicubic  & \shortstack{39.09\\0.9133} & \shortstack{39.78\\0.9215} & \shortstack{39.55\\0.9190} & \shortstack{40.67\\0.9315} & \shortstack{\textbf{40.71}\\\textbf{0.9319}} & \shortstack{40.64\\0.9311} & \shortstack{40.59\\0.9306} & \shortstack{40.66\\0.9313} & \shortstack{40.48\\0.9290}\\
  &    &    &  Bicubic + noise  & \shortstack{24.81\\0.4985} & \shortstack{32.98\\0.8856} & \shortstack{32.26\\0.8582} & \shortstack{34.65\\0.8908} & \shortstack{33.17\\0.8744} & \shortstack{34.81\\0.8951} & \shortstack{34.61\\0.8914} & \shortstack{\textbf{34.90}\\\textbf{0.8958}} & \shortstack{33.68\\0.8839}\\
\addlinespace[2pt]
  &  \multirow{2}{*}{Medium}  &  \multirow{2}{*}{74 / 22,137}  &  Bicubic  & \shortstack{30.63\\0.8219} & \shortstack{32.00\\0.8483} & \shortstack{31.80\\0.8452} & \shortstack{33.21\\0.8832} & \shortstack{\textbf{33.65}\\\textbf{0.8945}} & \shortstack{33.51\\0.8928} & \shortstack{33.14\\0.8837} & \shortstack{33.29\\0.8872} & \shortstack{33.12\\0.8844}\\
  &    &    &  Bicubic + noise  & \shortstack{23.34\\0.5114} & \shortstack{28.24\\0.7956} & \shortstack{28.02\\0.7787} & \shortstack{29.34\\0.8199} & \shortstack{29.48\\\textbf{0.8263}} & \shortstack{\textbf{29.61}\\0.8251} & \shortstack{29.43\\0.8194} & \shortstack{29.60\\0.8238} & \shortstack{28.95\\0.8069}\\
\addlinespace[2pt]
  &  \multirow{2}{*}{High}  &  \multirow{2}{*}{74 / 16,038}  &  Bicubic  & \shortstack{25.01\\0.7141} & \shortstack{26.87\\0.7722} & \shortstack{26.71\\0.7669} & \shortstack{27.60\\0.8095} & \shortstack{\textbf{27.98}\\\textbf{0.8246}} & \shortstack{\textbf{27.98}\\0.8234} & \shortstack{27.74\\0.8126} & \shortstack{27.59\\0.8091} & \shortstack{27.58\\0.8103}\\
  &    &    &  Bicubic + noise  & \shortstack{21.62\\0.4739} & \shortstack{25.18\\0.7096} & \shortstack{24.82\\0.6981} & \shortstack{25.78\\0.7360} & \shortstack{\textbf{26.00}\\\textbf{0.7492}} & \shortstack{25.75\\0.7377} & \shortstack{25.98\\0.7412} & \shortstack{25.90\\0.7392} & \shortstack{25.50\\0.7244}\\
\addlinespace[5pt]
\multirow{6}{*}{\shortstack{Temporal-noise\\proxy}}  &  \multirow{2}{*}{Low}  &  \multirow{2}{*}{74 / 17,662}  &  Bicubic  & \shortstack{36.75\\0.8953} & \shortstack{37.86\\0.9124} & \shortstack{37.62\\0.9097} & \shortstack{38.51\\0.9202} & \shortstack{\textbf{38.59}\\\textbf{0.9213}} & \shortstack{38.51\\0.9200} & \shortstack{38.38\\0.9182} & \shortstack{38.48\\0.9196} & \shortstack{38.36\\0.9180}\\
  &    &    &  Bicubic + noise  & \shortstack{24.49\\0.4907} & \shortstack{32.25\\0.8810} & \shortstack{31.54\\0.8531} & \shortstack{33.58\\0.8845} & \shortstack{32.33\\0.8712} & \shortstack{33.75\\0.8880} & \shortstack{33.55\\0.8853} & \shortstack{\textbf{33.86}\\\textbf{0.8901}} & \shortstack{32.65\\0.8772}\\
\addlinespace[2pt]
  &  \multirow{2}{*}{Medium}  &  \multirow{2}{*}{74 / 21,313}  &  Bicubic  & \shortstack{32.85\\0.8579} & \shortstack{34.27\\0.8849} & \shortstack{34.07\\0.8821} & \shortstack{35.36\\0.9041} & \shortstack{\textbf{35.56}\\\textbf{0.9073}} & \shortstack{35.41\\0.9045} & \shortstack{35.16\\0.9003} & \shortstack{35.28\\0.9023} & \shortstack{35.09\\0.8998}\\
  &    &    &  Bicubic + noise  & \shortstack{23.85\\0.5038} & \shortstack{29.51\\0.8280} & \shortstack{29.24\\0.8097} & \shortstack{30.68\\0.8417} & \shortstack{30.64\\0.8430} & \shortstack{30.89\\0.8449} & \shortstack{30.79\\0.8424} & \shortstack{\textbf{30.99}\\\textbf{0.8481}} & \shortstack{30.34\\0.8331}\\
\addlinespace[2pt]
  &  \multirow{2}{*}{High}  &  \multirow{2}{*}{74 / 15,869}  &  Bicubic  & \shortstack{24.05\\0.6789} & \shortstack{25.42\\0.7278} & \shortstack{25.27\\0.7222} & \shortstack{26.57\\0.7902} & \shortstack{27.27\\0.8161} & \shortstack{\textbf{27.31}\\\textbf{0.8168}} & \shortstack{26.96\\0.8005} & \shortstack{26.82\\0.7984} & \shortstack{26.79\\0.7984}\\
  &    &    &  Bicubic + noise  & \shortstack{21.17\\0.4932} & \shortstack{23.95\\0.6647} & \shortstack{23.68\\0.6563} & \shortstack{24.81\\0.7084} & \shortstack{\textbf{25.11}\\\textbf{0.7265}} & \shortstack{24.84\\0.7137} & \shortstack{24.96\\0.7117} & \shortstack{24.81\\0.7076} & \shortstack{24.45\\0.6911}\\
\addlinespace[5pt]
\multirow{8}{*}{\shortstack{Displacement\\(LR pixels)}}  &  \multirow{2}{*}{$[0,0.5)$}  &  \multirow{2}{*}{148 / 39,412}  &  Bicubic  & \shortstack{33.87\\0.8564} & \shortstack{35.16\\0.8830} & \shortstack{34.94\\0.8800} & \shortstack{36.08\\0.9013} & \shortstack{\textbf{36.23}\\\textbf{0.9044}} & \shortstack{36.08\\0.9012} & \shortstack{35.88\\0.8966} & \shortstack{36.01\\0.8995} & \shortstack{35.85\\0.8969}\\
  &    &    &  Bicubic + noise  & \shortstack{23.80\\0.4935} & \shortstack{30.20\\0.8385} & \shortstack{29.74\\0.8161} & \shortstack{31.42\\0.8493} & \shortstack{30.84\\0.8445} & \shortstack{31.62\\0.8530} & \shortstack{31.46\\0.8487} & \shortstack{\textbf{31.70}\\\textbf{0.8545}} & \shortstack{30.78\\0.8383}\\
\addlinespace[2pt]
  &  \multirow{2}{*}{$[0.5,1)$}  &  \multirow{2}{*}{29 / 8,444}  &  Bicubic  & \shortstack{27.15\\0.7386} & \shortstack{28.45\\0.7707} & \shortstack{28.29\\0.7669} & \shortstack{29.80\\0.8334} & \shortstack{\textbf{30.41}\\\textbf{0.8528}} & \shortstack{30.38\\0.8524} & \shortstack{29.95\\0.8374} & \shortstack{29.99\\0.8402} & \shortstack{29.82\\0.8359}\\
  &    &    &  Bicubic + noise  & \shortstack{22.56\\0.4998} & \shortstack{26.19\\0.7057} & \shortstack{25.88\\0.6954} & \shortstack{27.28\\0.7522} & \shortstack{\textbf{27.53}\\\textbf{0.7649}} & \shortstack{27.22\\0.7501} & \shortstack{27.29\\0.7501} & \shortstack{27.34\\0.7521} & \shortstack{26.89\\0.7358}\\
\addlinespace[2pt]
  &  \multirow{2}{*}{$[1,2)$}  &  \multirow{2}{*}{15 / 4,397}  &  Bicubic  & \shortstack{24.27\\0.6828} & \shortstack{25.36\\0.7150} & \shortstack{25.24\\0.7093} & \shortstack{26.67\\0.7845} & \shortstack{27.71\\0.8240} & \shortstack{\textbf{27.81}\\\textbf{0.8278}} & \shortstack{27.28\\0.8066} & \shortstack{27.19\\0.8046} & \shortstack{27.19\\0.8069}\\
  &    &    &  Bicubic + noise  & \shortstack{21.46\\0.5179} & \shortstack{23.95\\0.6512} & \shortstack{23.71\\0.6486} & \shortstack{25.05\\0.7100} & \shortstack{\textbf{25.47}\\\textbf{0.7362}} & \shortstack{25.26\\0.7268} & \shortstack{25.15\\0.7149} & \shortstack{25.08\\0.7101} & \shortstack{24.74\\0.6982}\\
\addlinespace[2pt]
  &  \multirow{2}{*}{$[2,\infty)$}  &  \multirow{2}{*}{30 / 2,591}  &  Bicubic  & \shortstack{23.20\\0.7257} & \shortstack{25.17\\0.7990} & \shortstack{24.93\\0.7920} & \shortstack{25.00\\0.7921} & \shortstack{25.45\\0.8078} & \shortstack{26.02\\0.8241} & \shortstack{\textbf{26.22}\\\textbf{0.8306}} & \shortstack{25.05\\0.7941} & \shortstack{25.56\\0.8122}\\
  &    &    &  Bicubic + noise  & \shortstack{20.93\\0.4956} & \shortstack{23.88\\0.7279} & \shortstack{23.53\\0.7138} & \shortstack{23.81\\0.7174} & \shortstack{24.14\\0.7338} & \shortstack{23.80\\0.7208} & \shortstack{\textbf{24.77}\\\textbf{0.7564}} & \shortstack{23.87\\0.7236} & \shortstack{24.03\\0.7300}\\
\bottomrule[1.2pt]
\end{tabular*}
\endgroup
\par\smallskip
\begin{minipage}{\linewidth}
Entries in (a) report PSNR (dB) / SSIM; cells in (b) list PSNR above SSIM. \textbf{Bold} marks the best PSNR and SSIM within each setting in (a) and each row in (b), including displayed ties. Both settings share the original-data groups. -- denotes unavailable measurements.
\end{minipage}
\endgroup
\vspace{-8pt}
\end{table*}

RGB implementations receive grayscale inputs replicated across three channels. For EDVR and BasicVSR++, the first output channel is used as the grayscale reconstruction. Method-specific input and temporal adaptations are detailed in Supplementary Section~S5.

PSNR and SSIM are computed after a 10-pixel border crop and averaged over 54,844 center frames from 222 test sequences. Group scores use the same frame-wise averaging, and neighboring-frame indices are clamped at sequence boundaries.

Latency is measured on an NVIDIA RTX 4090 at $160\times128$ LR resolution and batch size 1, excluding data loading and transfer. All baselines use FP32. FLOPs are estimated at $64\times64$ LR resolution for complete forward passes, including custom operators, with one multiply--accumulate counted as two operations.

Training configurations, adapted implementations, checkpoints, noise-generation code, and per-frame results will accompany the benchmark release.

\subsection{Benchmark Results}
Table~\ref{tab:sota_comparison}(a) compares seven learned baselines, the CGMamba reference model, and bicubic interpolation for $\times 4$ infrared super-resolution on IR275K. The learned baselines comprise two single-frame methods and five multi-frame methods.

Overall rankings differed between degradation settings. EDVR led under bicubic degradation, reaching 34.14~dB in PSNR, while MSDTGP led under mixed sensor noise with 30.13~dB. BasicVSR++ ranked second in both settings.

\begin{figure*}[!t]
\centering
\includegraphics[width=0.90\textwidth]{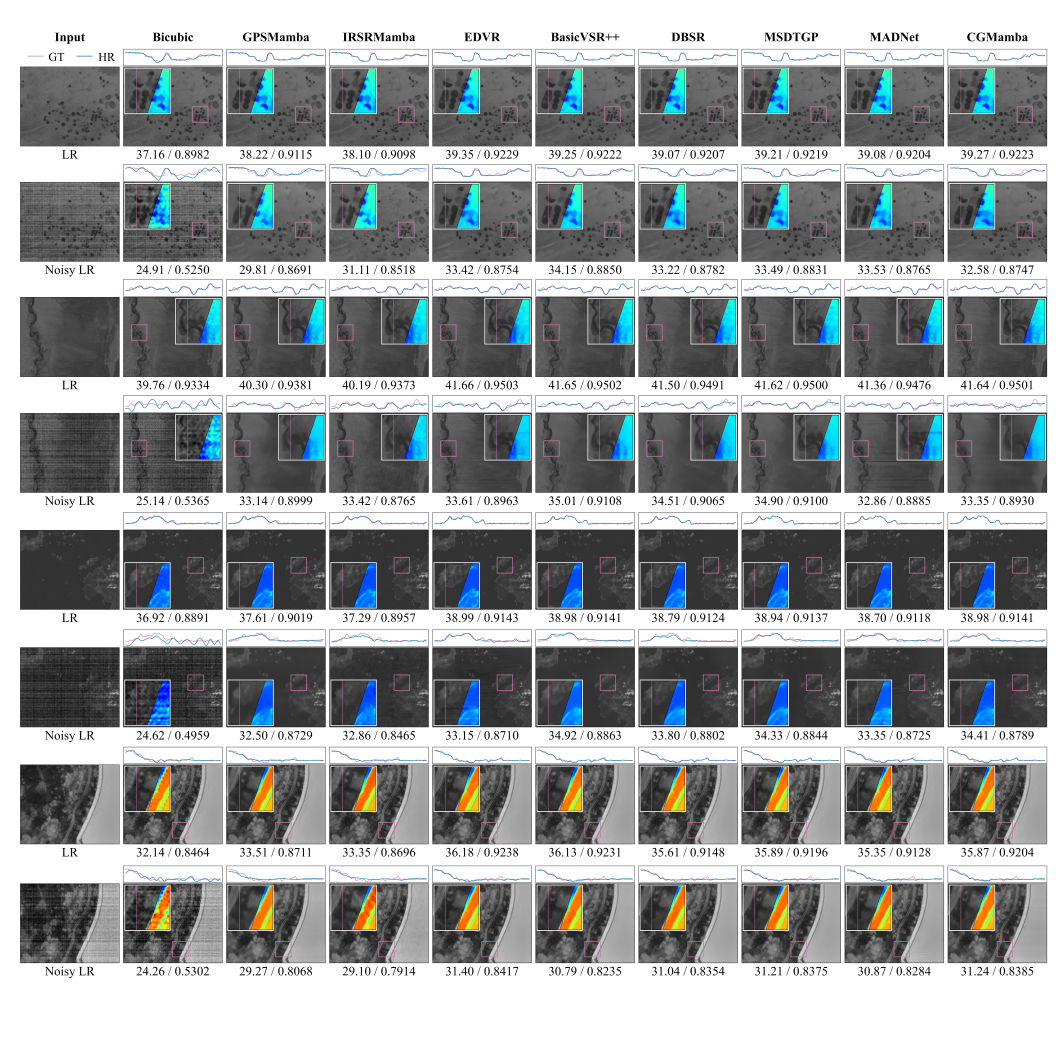}
\caption{Qualitative $\times 4$ comparison on four IR275K test scenes. Within each scene, upper and lower rows use bicubic LR inputs without and with mixed sensor noise, respectively. Columns show the input, Bicubic interpolation, two single-frame baselines (GPSMamba and IRSRMamba), five multi-frame baselines, and CGMamba. Insets show the same magnified region in grayscale and pseudocolor. Pink (GT) and blue (reconstruction) solid profiles above each image trace normalized intensities along the marked vertical line. Values below each reconstruction are full-frame PSNR (dB) / SSIM after a 10-pixel border crop.}
\label{fig:comparison}
\end{figure*}

Fig.~\ref{fig:comparison} complements these aggregate scores with paired reconstructions of the same four test frames. The matched regions and GT-referenced profiles allow direct comparison of local thermal boundaries and intensity transitions under both degradation settings.

\subsection{Performance Across Imaging Conditions}
\label{sec:condition_performance}

We evaluate reconstruction performance across source, contrast, temporal-noise-proxy, and inter-frame displacement groups (Table~\ref{tab:condition_performance}(b)). The descriptors follow the definitions in Section~\ref{sec:split_characterization}. Contrast and temporal-proxy groups follow test-sequence tertiles, yielding 74 sequences per group. Contrast boundaries are approximately 0.052345 and 0.181039; temporal-proxy boundaries are approximately 0.025101 and 0.110265. For boundaries $b_1<b_2$, the low, medium, and high groups correspond to $x<b_1$, $b_1\leq x<b_2$, and $x\geq b_2$, respectively. Both descriptors use population standard deviations on intensities normalized to $[0,1]$. Sequences are grouped into $[0,0.5)$, $[0.5,1)$, $[1,2)$, and $[2,\infty)$ LR-pixel displacement intervals.

Source-specific rankings differed from the overall ranking. Under mixed sensor noise, MSDTGP led on SatVideoIRSDT, DBSR on UAV-TSR++, and EDVR on SIATD. Although MSDTGP exceeded EDVR by 0.54~dB in overall PSNR, this advantage was concentrated in SatVideoIRSDT, where it led by 1.41~dB. It trailed EDVR by 0.25~dB on UAV-TSR++ and 0.08~dB on SIATD.

DBSR led the high-displacement UAV subset under both settings, despite not leading either overall ranking. Across the 30 sequences with displacement of at least 2~LR pixels, it exceeded EDVR in PSNR by 0.77~dB under bicubic degradation and 0.63~dB under mixed sensor noise. DBSR also achieved the highest SSIM in both settings.

Method rankings also changed across the contrast and temporal-noise-proxy groups. Under mixed sensor noise, MSDTGP led the low-contrast and low-temporal-proxy groups, whereas EDVR led the high-contrast and high-temporal-proxy groups. Source composition covaried with these descriptors: the low-contrast group contained only SatVideoIRSDT sequences, and the largest-displacement group contained only UAV-TSR++ sequences. The grouped results thus characterize combined acquisition conditions rather than isolate the effect of an individual descriptor. Comparisons across degradation settings use separately trained models.


\subsection{Ablation Study}
\label{sec:controlled_ablation}

\subsubsection{Temporal Information and Reconstruction Design}

Table~\ref{tab:ablation_structure} examines neighboring-frame input, center conditioning, feature aggregation, and the center-image residual. Center Q uses the center-frame feature at each group entrance; self Q uses each branch's own feature. The original aggregation head combines center-conditioned convolutional pre-fusion with temporal aggregation. The alternatives use concatenation followed by convolutions, or temporal attention with a center-feature query. Variant (b) repeats the same center frame three times, including its noise realization in the mixed-noise setting. Variant (g) omits the bilinearly upsampled center-image residual; feature-level center residuals are retained in all variants.

\Needspace{7\baselineskip}
All variants share the backbone width, depth, loss, and training protocol in Section~\ref{sec:experimental_setup}. Parameter counts range from 7.4345M to 7.4434M, differing from CGMamba by at most 0.1201\%. The paired comparisons (a, b), (a, c), (d, e), and (a, g) have identical parameter counts. Full-forward FLOP estimates are reported in Supplementary Table~S5.

\begin{wideablation}
\caption{Controlled Architectural Ablation on IR275K Under Bicubic and Mixed-Noise Degradation}
\label{tab:ablation_structure}
\centering
\begingroup
\small
\renewcommand{\arraystretch}{1.15}
\setlength{\tabcolsep}{3pt}
\begin{tabular*}{\textwidth}{@{\extracolsep{\fill}}cccccccc@{}}
\toprule[1.2pt]
\textbf{Variant} & \textbf{Input} & \textbf{CGCM Q} & \textbf{Aggregation} & \shortstack{\textbf{Image}\\\textbf{residual}} & \shortstack{\textbf{Params}\\\textbf{(M)}} & \shortstack{\textbf{Bicubic}\\\textbf{PSNR / SSIM}$\uparrow$} & \shortstack{\textbf{Bicubic + noise}\\\textbf{PSNR / SSIM}$\uparrow$}\\
\midrule[0.6pt]
(a) & Adjacent & Center & Original & Yes & 7.434476 & 33.83250\,/\,0.876345 & \textbf{29.91377}\,/\,\textbf{0.816922}\\
(b) & Center $\times3$ & Center & Original & Yes & 7.434476 & 32.78978\,/\,0.846167 & 28.96765\,/\,0.789901\\
(c) & Adjacent & Self & Original & Yes & 7.434476 & 33.67351\,/\,0.871872 & 29.63320\,/\,0.810946\\
(d) & Adjacent & Center & Concat conv. & Yes & 7.434508 & \textbf{33.83809}\,/\,\textbf{0.876805} & 29.73384\,/\,0.814464\\
(e) & Adjacent & Self & Concat conv. & Yes & 7.434508 & 33.57607\,/\,0.869600 & 29.60713\,/\,0.812908\\
(f) & Adjacent & Self & Temporal attn. & Yes & 7.443404 & 33.47114\,/\,0.864608 & 29.61227\,/\,0.809319\\
(g) & Adjacent & Center & Original & No & 7.434476 & 33.67347\,/\,0.871912 & 29.50440\,/\,0.804338\\
\bottomrule[1.2pt]
\end{tabular*}
\par\smallskip
\begin{minipage}{\linewidth}
\footnotesize
All variants use three grayscale input branches and $\times4$ center-frame reconstruction. Original denotes the CGMamba aggregation head. Each cell reports PSNR (dB) / SSIM; bold marks the best value of each metric within each setting. Each configuration and degradation setting uses one training run.
\end{minipage}
\endgroup
\end{wideablation}

Real neighboring frames improved PSNR over center-frame repetition by 1.04~dB under bicubic degradation and 0.95~dB under mixed sensor noise. Center conditioning also improved performance within both aggregation heads: gains over self conditioning ranged from 0.16 to 0.26~dB under bicubic degradation and from 0.13 to 0.28~dB under mixed sensor noise. SSIM also improved in all four comparisons. The consistent direction across heads and degradation settings supports center-conditioned interaction beyond the benefit of supplying additional input frames.

With center-guided CGCM retained, the original aggregation head achieved similar bicubic PSNR to concatenation and a 0.18~dB advantage under mixed sensor noise. The center-image residual improved PSNR by 0.16~dB under bicubic degradation and 0.41~dB under mixed sensor noise. Both reconstruction choices therefore provided larger benefits in the mixed-noise setting.

Under bicubic degradation, the value of neighboring frames varied across acquisition conditions. Relative to center-frame repetition, CGMamba gained 0.98, 0.17, and 1.20~dB on SatVideoIRSDT, UAV-TSR++, and SIATD, respectively. In the group with displacement of at least 2~LR pixels, the gain was only 0.07~dB, with higher sequence-mean PSNR in 13 of 30 sequences. Thus, the aggregate multi-frame gain coexisted with limited and uneven benefit on the high-displacement UAV subset. Source- and displacement-specific results are given in Supplementary Table~S4.

\section{Conclusion}

We introduced IR275K, a benchmark that unifies 594 infrared video sequences and 275,196 frames from three public sources under a common $\times4$ reconstruction protocol. Its sequence-level partitions and paired single-frame and multi-frame evaluations provide a shared basis for reproducible comparison under bicubic degradation and synthetic mixed sensor noise. Source- and condition-specific analyses reveal performance differences hidden by aggregate scores. Method rankings vary across acquisition conditions, motivating evaluation beyond a single overall ranking.

CGMamba complements this benchmark with a compact, center-guided state-space architecture for implicit multi-frame reconstruction. Parameter-matched controls demonstrate PSNR gains of 1.04~dB and 0.95~dB from real neighboring frames under the two degradation settings. Center-frame conditioning also improves reconstruction across both aggregation heads and settings, supporting reference-guided feature interaction as a useful architectural choice. Together, IR275K and CGMamba provide both a common evaluation resource and a concrete framework for investigating and improving temporal information use in infrared reconstruction.

Further extensions could incorporate sensor-calibrated degradation models, longer temporal contexts, and downstream remote-sensing tasks.

\begin{IEEEbiography}[{\includegraphics[width=1in,height=1.25in,clip,keepaspectratio]{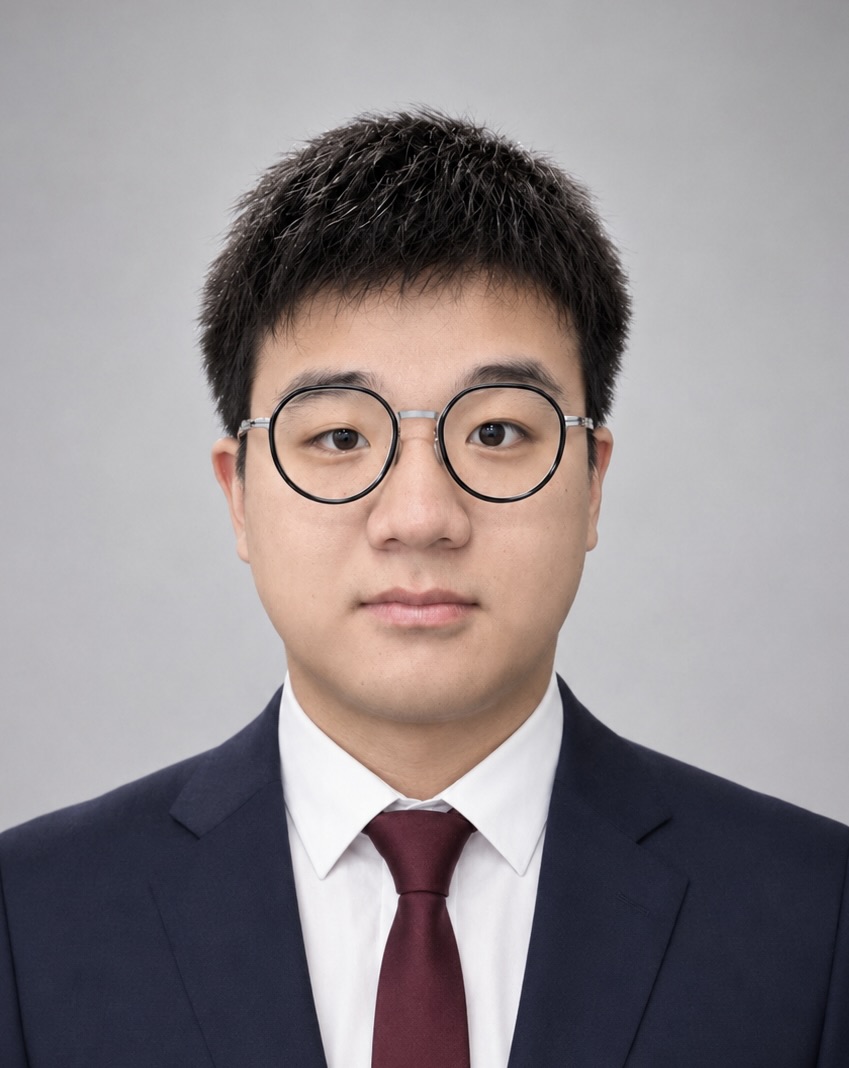}}]{Jie Deng}
Jie Deng received the B.Eng. degree from Hunan University of Technology, Zhuzhou, China. He is currently pursuing the M.Eng. degree with the Hangzhou Institute for Advanced Study, Hangzhou, China, and the University of Chinese Academy of Sciences, Beijing, China. His current research interests include computational imaging, artificial intelligence-based imaging, infrared multi-frame super-resolution, and remote sensing image preprocessing.
\end{IEEEbiography}
\vspace{-1.0em}
\begin{IEEEbiography}[{\includegraphics[width=1in,height=1.25in,clip,keepaspectratio]{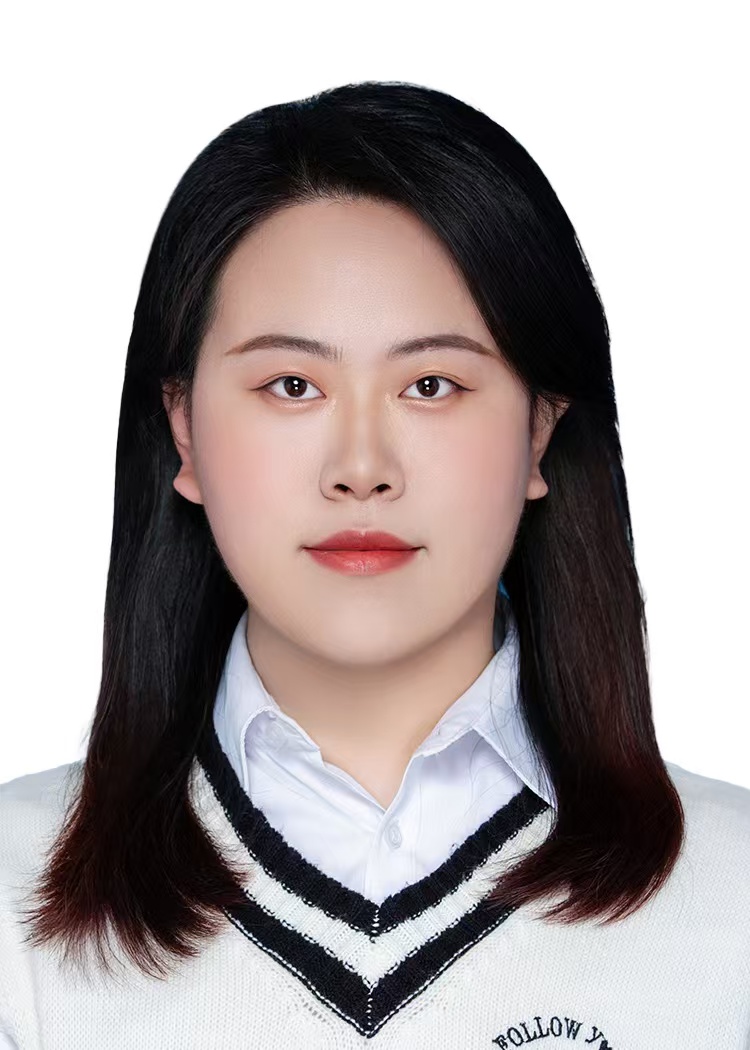}}]{Heyang Wang}
Heyang Wang (Student Member, IEEE) received the B.Sc. degree in mathematics and applied mathematics from Harbin Normal University, Harbin, China, in 2022. She is currently pursuing the M.Sc. degree in physical electronics with the Hangzhou Institute for Advanced Study, Hangzhou, China, and the University of Chinese Academy of Sciences, Beijing, China. Her current research interests include computer vision and remote sensing image processing.
\end{IEEEbiography}
\vspace{-1.0em}
\begin{IEEEbiography}[{\includegraphics[width=1in,height=1.25in,clip,keepaspectratio]{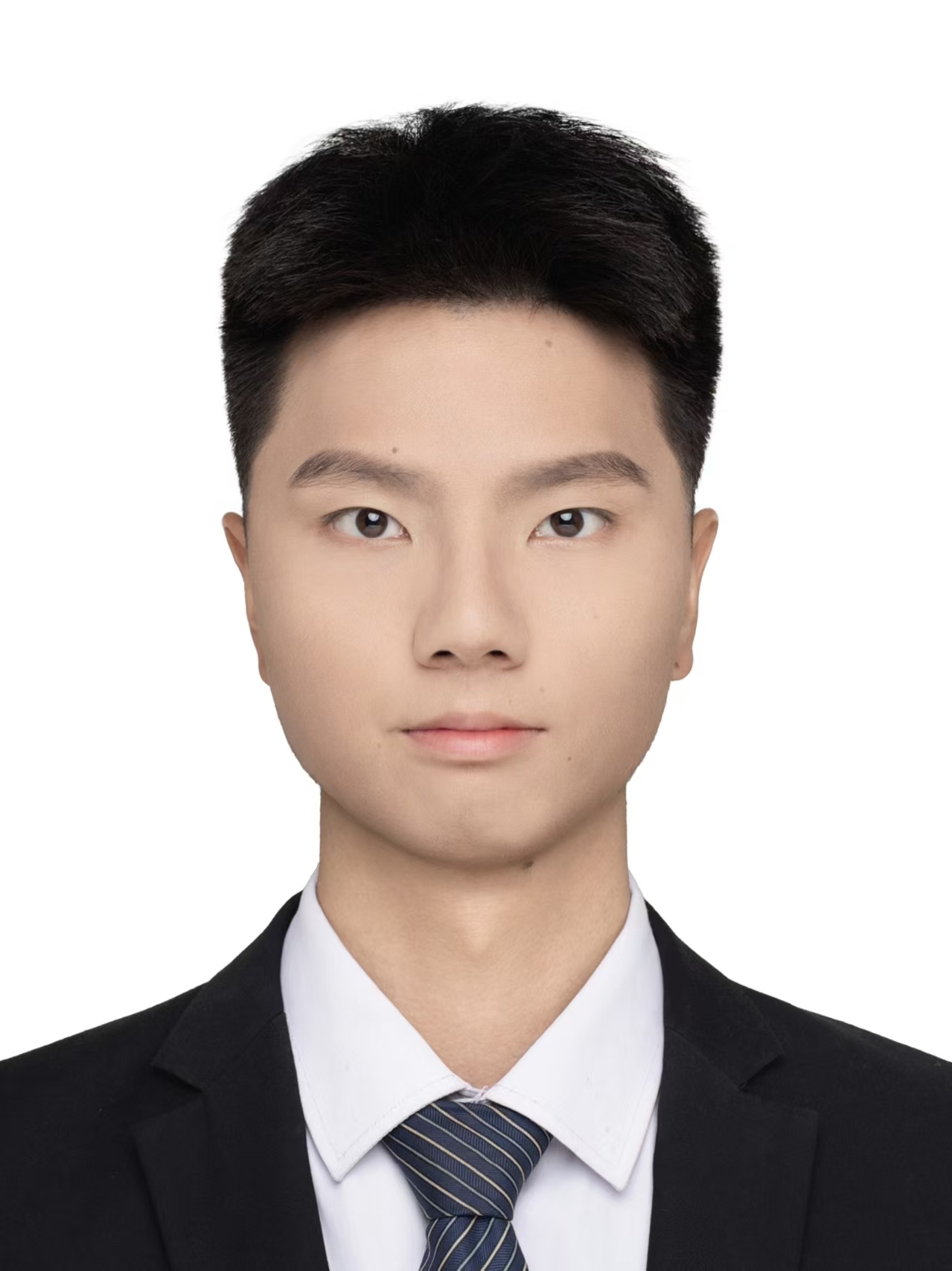}}]{Changxin Wang}
Changxin Wang received the B.Eng. degree from China University of Mining and Technology, Xuzhou, China. He is currently pursuing the M.Eng. degree with the Hangzhou Institute for Advanced Study, Hangzhou, China, and the University of Chinese Academy of Sciences, Beijing, China. His current research interests include visual SLAM, LiDAR SLAM, multi-sensor fusion, and robust 3-D perception and localization in degraded environments.
\end{IEEEbiography}
\vspace{-1.0em}
\begin{IEEEbiography}[{\includegraphics[width=1in,height=1.25in,clip,keepaspectratio]{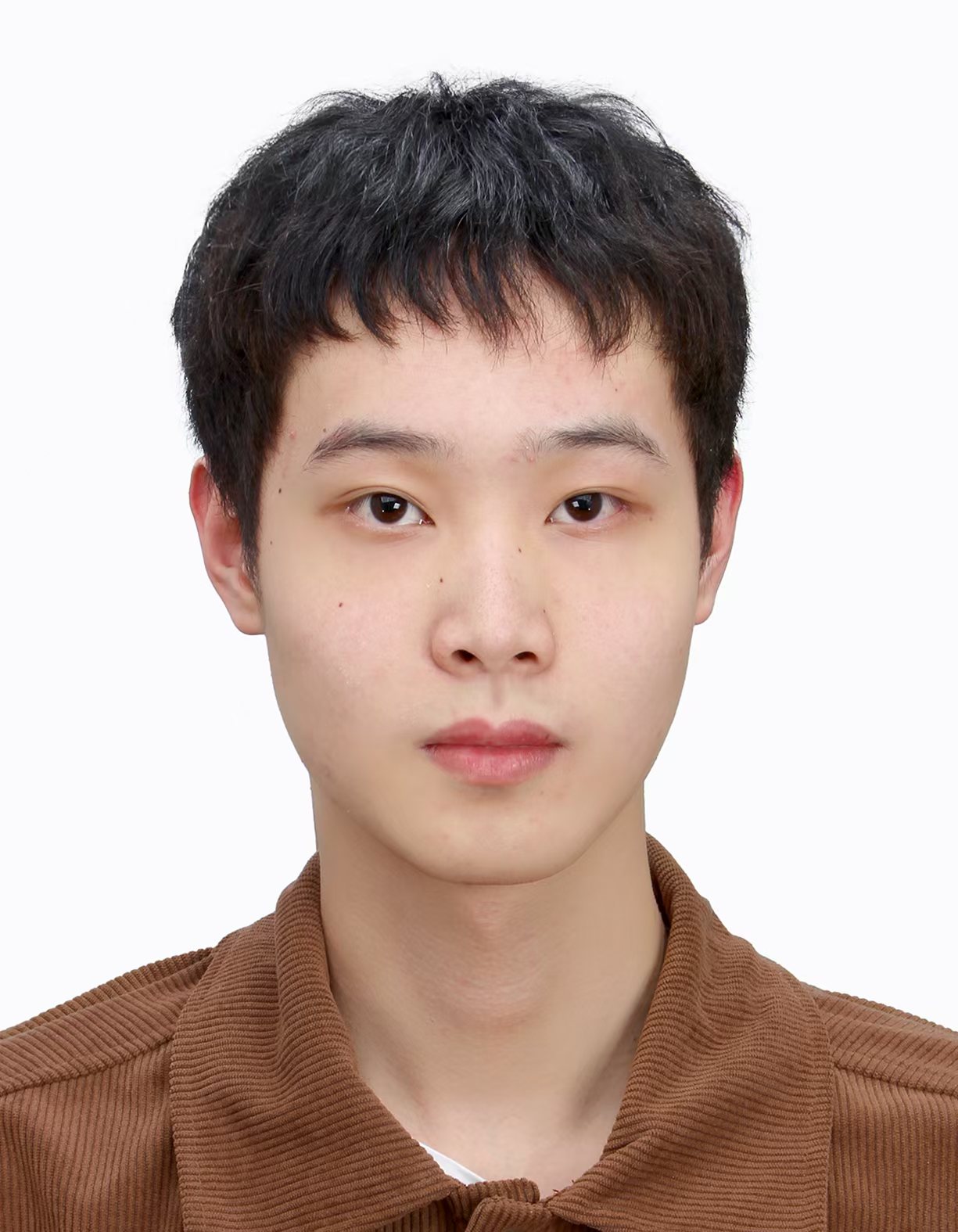}}]{Junkai Shen}
Junkai Shen is currently pursuing the M.Eng. degree in physical electronics with the Hangzhou Institute for Advanced Study, Hangzhou, China, and the University of Chinese Academy of Sciences, Beijing, China. His current research interests include visual SLAM and LiDAR SLAM.
\end{IEEEbiography}
\vspace{-1.0em}
\begin{IEEEbiography}[{\includegraphics[width=1in,height=1.25in,clip,keepaspectratio]{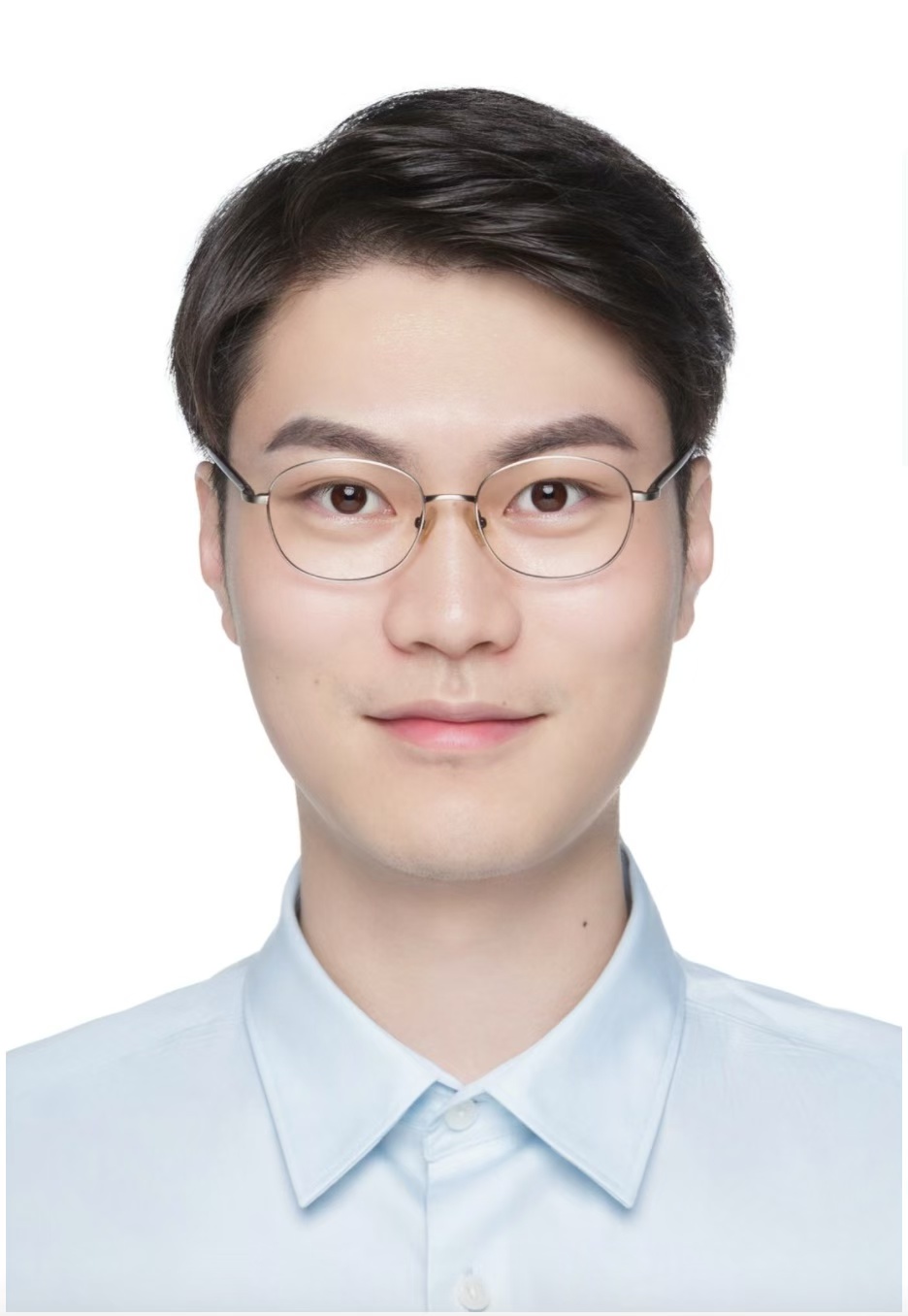}}]{Hongyi Chen}
Hongyi Chen received the Ph.D. degree from the School of Engineering, University of Edinburgh, Edinburgh, U.K., in 2021. He is currently a Postdoctoral Researcher with the Hangzhou Institute for Advanced Study, Hangzhou, China, and the University of Chinese Academy of Sciences, Beijing, China. His research interests include computer vision, remote sensing image processing, physics-informed machine learning, and edge computing.
\end{IEEEbiography}
\vspace{-1.0em}
\begin{IEEEbiography}[{\includegraphics[width=1in,height=1.25in,clip,keepaspectratio]{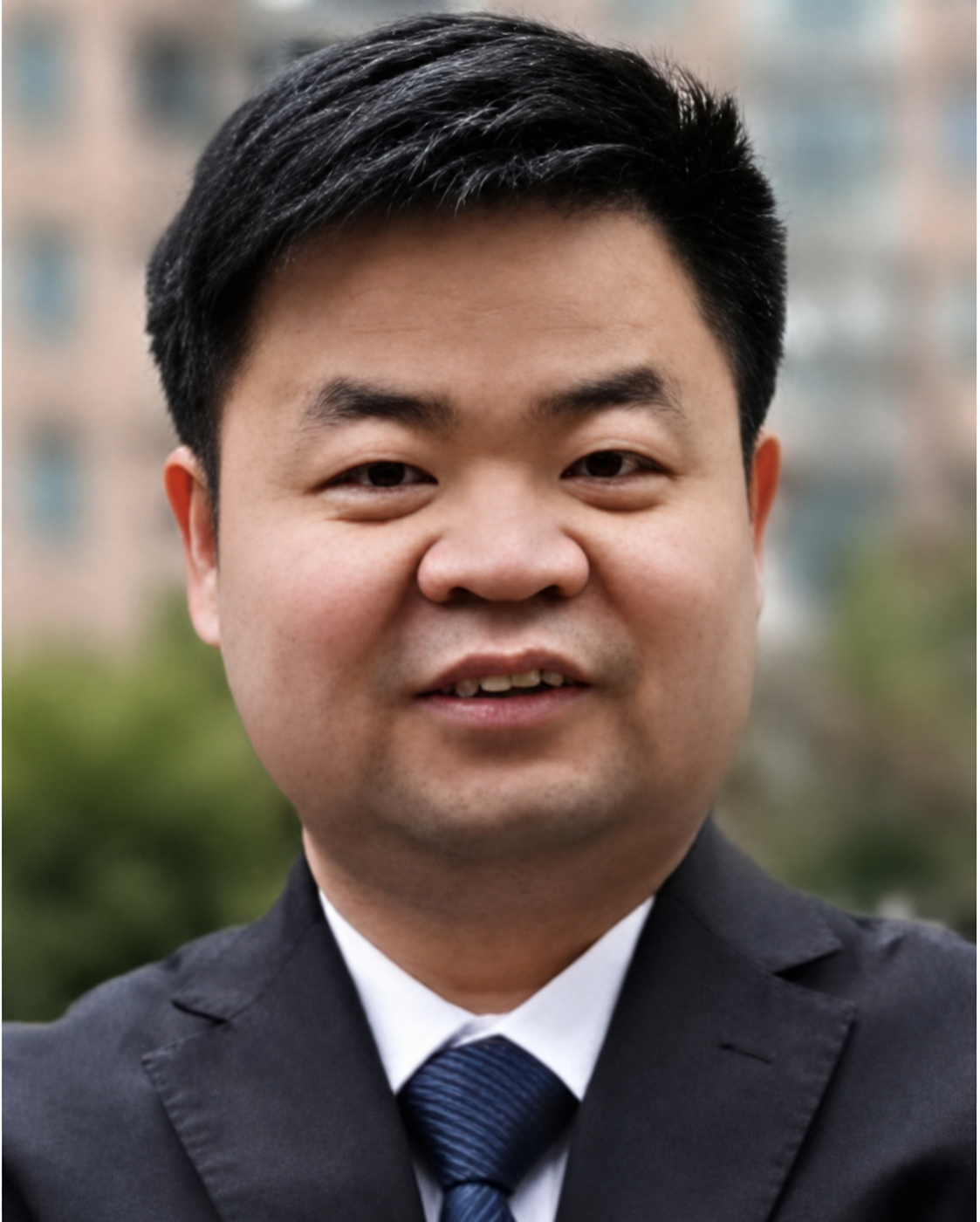}}]{Zhiping He}
Zhiping He received the M.S. degree in optoelectronic engineering from Zhejiang University, Zhejiang, China, in 2003, and the Ph.D. degree from the Shanghai Institute of Technical Physics, Chinese Academy of Sciences, Shanghai, China, in 2009. He has long been engaged in research on space optics and optoelectronic technology, with a focus on key technologies for space spectral sensing and active-passive composite optical systems. The instruments developed by his team have been successfully applied to China's Chang'e-3, Chang'e-4, Chang'e-5, and Mars exploration projects.
\end{IEEEbiography}
\vspace{-1.0em}
\begin{IEEEbiography}[{\includegraphics[width=1in,height=1.25in,clip,keepaspectratio]{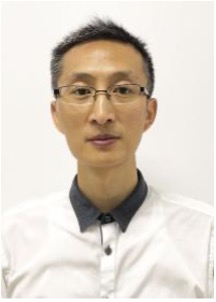}}]{Hongxing Qi}
Hongxing Qi received the Ph.D. degree in physical electronics from the Shanghai Institute of Technical Physics, Chinese Academy of Sciences, Shanghai, China, in 2007. He is currently a Researcher and Doctoral Supervisor with the Hangzhou Institute for Advanced Study, Hangzhou, China, and the University of Chinese Academy of Sciences, Beijing, China. His current research interests include optoelectronic technology and devices.
\end{IEEEbiography}
\vspace{0 em}
\begin{IEEEbiography}[{\includegraphics[width=1in,height=1.25in,clip,keepaspectratio]{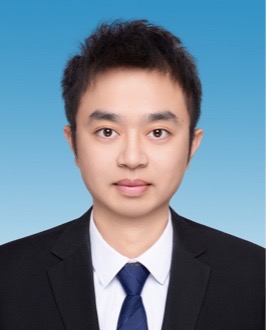}}]{Xudong Zhang}
Xudong Zhang received the Ph.D. degree from the Shanghai Institute of Technical Physics, Chinese Academy of Sciences, Shanghai, China, in 2019. He is currently a Distinguished Research Fellow with the Hangzhou Institute for Advanced Study, Hangzhou, China, and the University of Chinese Academy of Sciences, Beijing, China. His research interests include spatial intelligence, 3-D reconstruction, and artificial intelligence algorithms for edge devices, such as robots, satellite payloads, and unmanned aerial vehicles.
\end{IEEEbiography}
\vfill


\begin{thebibliography}{99}

\bibitem{ref46}
Q. Jiang, L. Zheng, Y. Zhou, H. Liu, Q. Kong, Y. Zhang, and B. Chen, ``Efficient on-orbit remote sensing imagery processing via satellite edge computing resource scheduling optimization,''
\textit{IEEE Trans. Geosci. Remote Sens.}, vol.~63, pp.~1--19, 2025, doi:~10.1109/TGRS.2025.3528015.

\bibitem{ref50}
Q. Fang, H. Wang, G.-H. He, Z.-Y. Zhou, F.-W. Cao, and Q.-H. Song, ``Research on spaceborne full-waveform LiDAR filtering processing technology based on bat algorithm Gaussian sharpening,''
\textit{J. Infrared Millim. Waves}, vol.~44, no.~6, pp.~856--864, 2025, doi:~10.11972/j.issn.1001-9014.2025.06.004.

\bibitem{ref51}
J. Yang, Y. Ma, W.-B. Yu, S.-H. Li, J. Yu, Q.-Y. Wang, and S. Li, ``Noise model of oceanic spaceborne photon counting LiDAR,''
\textit{J. Infrared Millim. Waves}, vol.~43, no.~3, pp.~393--398, 2024, doi:~10.11972/j.issn.1001-9014.2024.03.013.

\bibitem{ref52}
C.-T. Tan, W.-B. Yu, Y.-Y. Xiang, S.-H. Li, J. Yu, Q.-Y. Wang, and S. Li, ``Real-time denoising method for spaceborne photon counting laser ranging radar,''
\textit{J. Infrared Millim. Waves}, vol.~43, no.~2, pp.~241--253, 2024, doi:~10.11972/j.issn.1001-9014.2024.02.014.

\bibitem{ref33}
J. Li, X. Shi, B. Liu, R. Wang, Z. Wang, M. Wang, G. Lv, L. Yuan, Q. Ma, G. Yuan, \textit{et al.}, ``High-fidelity in-situ spectroscopy under thermal extremes enables cross-scale validation of the Chang'e-6 landing area,''
\textit{PhotoniX}, vol.~7, no.~1, p.~44, 2026, doi:~10.1186/s43074-026-00257-z.

\bibitem{ref24}
S. Nah, S. Baik, S. Hong, G. Moon, S. Son, R. Timofte, and K.~M. Lee, ``NTIRE 2019 challenge on video deblurring and super-resolution: Dataset and study,''
in \textit{Proc. IEEE/CVF Conf. Comput. Vis. Pattern Recognit. Workshops (CVPRW)}, 2019, pp.~1996--2005, doi:~10.1109/CVPRW.2019.00251.

\bibitem{ref25}
T. Xue, B. Chen, J. Wu, \textit{et al.}, ``Video enhancement with task-oriented flow,''
\textit{Int. J. Comput. Vis.}, vol.~127, no.~8, pp.~1106--1125, 2019.

\bibitem{ref26}
E. Agustsson and R. Timofte, ``NTIRE 2017 challenge on single image super-resolution: Dataset and study,''
in \textit{Proc. IEEE Conf. Comput. Vis. Pattern Recognit. Workshops (CVPRW)}, 2017, pp.~126--135.

\bibitem{ref47}
F. Bao, S. Jape, A. Schramka, J. Wang, T.~E. McGraw, and Z. Jacob, ``Why thermal images are blurry,''
\textit{Opt. Express}, vol.~32, no.~3, pp.~3852--3865, 2024.

\bibitem{li2026satvideodataset}
R. Li, Z. Li, N. Chen, \textit{et al.}, ``Infrared video satellite aerial moving target detection dataset and its evaluation,''
\textit{Journal of Image and Graphics}, vol.~31, no.~9, pp.~3321--3335, 2026, doi:~10.11834/jig.250536.

\bibitem{ref31}
R. Li, Y. Zeng, W. Sheng, \textit{et al.}, ``Infrared video satellite aerial moving target detection dataset,'' V2, Science Data Bank, 2025, doi:~\url{10.57760/sciencedb.j00240.00077}. [Online]. Available: \url{https://cstr.cn/31253.11.sciencedb.j00240.00077}. Accessed: Jul. 23, 2026. CSTR:~31253.11.sciencedb.j00240.00077.

\bibitem{ref32}
X. Sun, L. Guo, W. Zhang, \textit{et al.}, ``A dataset of semi-synthetic detection for small infrared moving targets under complex backgrounds,'' V3, Science Data Bank, 2024. [Online]. Available: \url{https://cstr.cn/31253.11.sciencedb.j00001.00231}. Accessed: Jul. 23, 2026. CSTR:~31253.11.sciencedb.j00001.00231.

\bibitem{ref30}
M. Li, C. Fu, Z. Lu, Z. Zhang, H. Zuo, and L. Yao, ``AnyTSR++: Prompt-oriented any-scale thermal super-resolution for unmanned aerial vehicle,''
\textit{IEEE Trans. Geosci. Remote Sens.}, vol.~63, pp.~1--15, 2025, doi:~10.1109/TGRS.2025.3630597.

\bibitem{Cai2024}
L. Cai, X. Dong, K. Zhou, and X. Cao, ``Exploring video denoising in thermal infrared imaging: Physics-inspired noise generator, dataset, and model,''
\textit{IEEE Trans. Image Process.}, vol.~33, pp.~3839--3854, 2024, doi:~10.1109/TIP.2024.3390404.

\bibitem{ref1}
C. Dong, C.~C. Loy, K. He, and X. Tang, ``Image super-resolution using deep convolutional networks,''
\textit{IEEE Trans. Pattern Anal. Mach. Intell.}, vol.~38, no.~2, pp.~295--307, 2016.

\bibitem{ref2}
B. Lim, S. Son, H. Kim, S. Nah, and K.~M. Lee, ``Enhanced deep residual networks for single image super-resolution,''
in \textit{Proc. IEEE Conf. Comput. Vis. Pattern Recognit. Workshops (CVPRW)}, 2017, pp.~136--144.

\bibitem{ref3}
Y. Zhang, Y. Tian, Y. Kong, B. Zhong, and Y. Fu, ``Residual dense network for image super-resolution,''
in \textit{Proc. IEEE Conf. Comput. Vis. Pattern Recognit. (CVPR)}, 2018, pp.~2472--2481.

\bibitem{ref4}
W.-S. Lai, J.-B. Huang, N. Ahuja, and M.-H. Yang, ``Deep Laplacian pyramid networks for fast and accurate super-resolution,''
in \textit{Proc. IEEE Conf. Comput. Vis. Pattern Recognit. (CVPR)}, 2017, pp.~624--632.

\bibitem{ref7}
J. Kim, J.~K. Lee, and K.~M. Lee, ``Accurate image super-resolution using very deep convolutional networks,''
in \textit{Proc. IEEE Conf. Comput. Vis. Pattern Recognit. (CVPR)}, 2016, pp.~1646--1654.

\bibitem{ref8}
Y. Zhang, K. Li, K. Li, L. Wang, B. Zhong, and Y. Fu, ``Image super-resolution using very deep residual channel attention networks,''
in \textit{Proc. Eur. Conf. Comput. Vis. (ECCV)}, 2018, pp.~286--301.

\bibitem{ref9}
T. Dai, J. Cai, Y. Zhang, S.-T. Xia, and L. Zhang, ``Second-order attention network for single image super-resolution,''
in \textit{Proc. IEEE/CVF Conf. Comput. Vis. Pattern Recognit. (CVPR)}, 2019, pp.~11065--11074.

\bibitem{ref6}
S. Lei and Z. Shi, ``Hybrid-scale self-similarity exploitation for remote sensing image super-resolution,''
\textit{IEEE Trans. Geosci. Remote Sens.}, vol.~60, pp.~1--10, 2022, doi:~10.1109/TGRS.2021.3069889.

\bibitem{ref5}
Y. Mei, Y. Fan, and Y. Zhou, ``Image super-resolution with non-local sparse attention,''
in \textit{Proc. IEEE/CVF Conf. Comput. Vis. Pattern Recognit. (CVPR)}, 2021, pp.~3517--3526.

\bibitem{ref10}
X. Wang, R. Girshick, A. Gupta, and K. He, ``Non-local neural networks,''
in \textit{Proc. IEEE Conf. Comput. Vis. Pattern Recognit. (CVPR)}, 2018, pp.~7794--7803.

\bibitem{ref11}
A. Vaswani \textit{et al.}, ``Attention is all you need,''
in \textit{Adv. Neural Inf. Process. Syst. (NeurIPS)}, vol.~30, 2017, pp.~5998--6008.

\bibitem{ref12}
Z. Liu, Y. Lin, Y. Cao, H. Hu, Y. Wei, Z. Zhang, S. Lin, and B. Guo, ``Swin Transformer: Hierarchical vision Transformer using shifted windows,''
in \textit{Proc. IEEE/CVF Int. Conf. Comput. Vis. (ICCV)}, 2021, pp.~10012--10022.

\bibitem{ref13}
S.~W. Zamir, A. Arora, S. Khan, M. Hayat, F.~S. Khan, and M.-H. Yang, ``Restormer: Efficient Transformer for high-resolution image restoration,''
in \textit{Proc. IEEE/CVF Conf. Comput. Vis. Pattern Recognit. (CVPR)}, 2022, pp.~5728--5739.

\bibitem{ref53}
J. Liang, J. Cao, G. Sun, K. Zhang, L. Van~Gool, and R. Timofte, ``SwinIR: Image restoration using Swin Transformer,''
in \textit{Proc. IEEE/CVF Int. Conf. Comput. Vis. Workshops (ICCVW)}, 2021, pp.~1833--1844.

\bibitem{ref14}
S. Lei, Z. Shi, and W. Mo, ``Transformer-based multistage enhancement for remote sensing image super-resolution,''
\textit{IEEE Trans. Geosci. Remote Sens.}, vol.~60, pp.~1--11, 2022, doi:~10.1109/TGRS.2021.3136190.

\bibitem{ref15}
X. Chen, X. Wang, J. Zhou, Y. Qiao, and C. Dong, ``Activating more pixels in image super-resolution Transformer,''
in \textit{Proc. IEEE/CVF Conf. Comput. Vis. Pattern Recognit. (CVPR)}, 2023, pp.~22367--22377.

\bibitem{ref16}
J.-F. Hu, T.-Z. Huang, L.-J. Deng, H.-X. Dou, D. Hong, and G. Vivone, ``Fusformer: A Transformer-based fusion network for hyperspectral image super-resolution,''
\textit{IEEE Geosci. Remote Sens. Lett.}, vol.~19, pp.~1--5, 2022, doi:~10.1109/LGRS.2022.3194257.

\bibitem{PortraitSR}
M. Wang, J. Leng, S. Li, C. Kuang, and L. Sun, ``PortraitSR: Artist-inspired prior learning for progressive face super-resolution,''
in \textit{Proc. AAAI Conf. Artif. Intell.}, vol.~40, no.~12, 2026, pp.~9984--9992, doi:~10.1609/aaai.v40i12.37964.

\bibitem{ref37}
Y. Huang, T. Miyazaki, X. Liu, and S. Omachi, ``GPSMamba: A global phase and spectral prompt-guided Mamba for infrared image super-resolution,''
\textit{arXiv preprint arXiv:2507.18998}, 2025, doi:~10.48550/arXiv.2507.18998.

\bibitem{ref38}
Y. Huang, T. Miyazaki, X. Liu, and S. Omachi, ``IRSRMamba: Infrared image super-resolution via Mamba-based wavelet transform feature modulation model,''
\textit{IEEE Trans. Geosci. Remote Sens.}, vol.~63, pp.~1--16, 2025, doi:~10.1109/TGRS.2025.3584385.

\bibitem{ref39}
X. Li, Z. Wang, Y. Zou, Z. Chen, J. Ma, Z. Jiang, L. Ma, and J. Liu, ``DifIISR: A diffusion model with gradient guidance for infrared image super-resolution,''
in \textit{Proc. IEEE/CVF Conf. Comput. Vis. Pattern Recognit. (CVPR)}, 2025, pp.~7534--7544.

\bibitem{li2026probing}
R. Li, W. An, Y. Wang, X. Ying, Y. Dai, L. Wang, M. Li, Y. Guo, and L. Liu, ``Probing deep into temporal profile makes the infrared small target detector much better,''
\textit{IEEE Trans. Pattern Anal. Mach. Intell.}, vol.~48, no.~8, pp.~10157--10175, 2026, doi:~10.1109/TPAMI.2026.3683258.

\bibitem{Li2026DHiF}
R. Li, C. Xiao, Q. Yin, W. An, N. Chen, X. Ying, M. Li, and Y. Wang, ``Dynamic high-frequency convolution for infrared small target detection,''
\textit{IEEE Trans. Circuits Syst. Video Technol.}, vol.~36, no.~6, pp.~7676--7680, 2026, doi:~10.1109/TCSVT.2026.3661285.

\bibitem{ref27}
J. Caballero, C. Ledig, A. Aitken, A. Acosta, J. Totz, Z. Wang, and W. Shi, ``Real-time video super-resolution with spatio-temporal networks and motion compensation,''
in \textit{Proc. IEEE Conf. Comput. Vis. Pattern Recognit. (CVPR)}, 2017, pp.~4778--4787.

\bibitem{ref41}
Y. Xiao, X. Su, Q. Yuan, D. Liu, H. Shen, and L. Zhang, ``Satellite video super-resolution via multiscale deformable convolution alignment and temporal grouping projection,''
\textit{IEEE Trans. Geosci. Remote Sens.}, vol.~60, pp.~1--19, 2022, doi:~10.1109/TGRS.2021.3107352.

\bibitem{ref40}
Y. Huang, X. Zhi, W. Chen, X. Liang, Z. Wang, and W. Zhang, ``DCUNet: Deformable convolutional UNet for multi-frame infrared small target super-resolution,''
in \textit{Proc. IEEE Int. Conf. Signal, Inf. Data Process. (ICSIDP)}, 2024, pp.~1--7, doi:~10.1109/ICSIDP62679.2024.10867882.

\bibitem{ref42}
D.~P. Tran, D.~D. Hung, and D. Kim, ``VSRM: A robust Mamba-based framework for video super-resolution,''
in \textit{Proc. IEEE/CVF Int. Conf. Comput. Vis. (ICCV)}, 2025, pp.~14711--14721.

\bibitem{ICNet}
J. Leng, J. Wang, X. Gao, B. Hu, J. Gan, and C. Gao, ``ICNet: Joint alignment and reconstruction via iterative collaboration for video super-resolution,''
in \textit{Proc. 30th ACM Int. Conf. Multimedia (ACM MM)}, 2022, pp.~6675--6684, doi:~10.1145/3503161.3547994.

\bibitem{ref43}
G. Bhat, M. Danelljan, F. Yu, L.~Van~Gool, and R. Timofte, ``Deep reparametrization of multi-frame super-resolution and denoising,''
in \textit{Proc. IEEE/CVF Int. Conf. Comput. Vis. (ICCV)}, 2021, pp.~2460--2470.

\bibitem{DDLBSR}
J. Leng, J. Wang, M. Mo, J. Gan, W. Lu, and X. Gao, ``Difficulty-guided variant degradation learning for blind image super-resolution,''
\textit{IEEE Trans. Neural Netw. Learn. Syst.}, vol.~36, no.~7, pp.~13080--13093, 2025, doi:~10.1109/TNNLS.2024.3462490.

\bibitem{ref17}
A. Gu and T. Dao, ``Mamba: Linear-time sequence modeling with selective state spaces,''
in \textit{Proc. 1st Conf. Lang. Model. (COLM)}, 2024.

\bibitem{ref22}
L. Zhu, B. Liao, Q. Zhang, X. Wang, W. Liu, and X. Wang, ``Vision Mamba: Efficient visual representation learning with bidirectional state space model,''
in \textit{Proc. 41st Int. Conf. Mach. Learn. (ICML)}, 2024, pp.~62429--62442.

\bibitem{ref18}
K. Jiang, M. Yang, Y. Xiao, J. Wu, G. Wang, X. Feng, and J. Jiang, ``Rep-Mamba: Re-parameterization in vision Mamba for lightweight remote sensing image super-resolution,''
\textit{IEEE Trans. Geosci. Remote Sens.}, vol.~63, pp.~1--12, 2025, doi:~10.1109/TGRS.2025.3597745.

\bibitem{ref19}
C. Li, Z. Pan, and D. Hong, ``Dynamic state-control modeling for generalized remote sensing image super-resolution,''
in \textit{Proc. IEEE/CVF Conf. Comput. Vis. Pattern Recognit. Workshops (CVPRW)}, 2025, pp.~3067--3075, doi:~10.1109/CVPRW67362.2025.00290.

\bibitem{ref20}
T. Wu, R. Zhao, M. Lv, Z. Jia, L. Li, M. Liu, X. Zhao, H. Ma, and G. Vivone, ``Efficient Mamba-attention network for remote sensing image super-resolution,''
\textit{IEEE Trans. Geosci. Remote Sens.}, vol.~63, pp.~1--14, 2025, doi:~10.1109/TGRS.2025.3578879.

\bibitem{ref21}
Y. Xiao, Q. Yuan, K. Jiang, \textit{et al.}, ``Frequency-assisted Mamba for remote sensing image super-resolution,''
\textit{IEEE Trans. Multimedia}, vol.~27, pp.~1783--1796, 2025, doi:~10.1109/TMM.2024.3521798.

\bibitem{ref34}
M. Li, C. Xiong, Z. Gao, and J. Ma, ``HAM: Hierarchical attention Mamba with spatial-frequency fusion for remote sensing image super-resolution,''
\textit{IEEE Trans. Geosci. Remote Sens.}, vol.~63, pp.~1--14, 2025, doi:~10.1109/TGRS.2025.3605861.

\bibitem{ref35}
X. Lei, W. Zhang, and W. Cao, ``DVMSR: Distillated vision Mamba for efficient super-resolution,''
in \textit{Proc. IEEE/CVF Conf. Comput. Vis. Pattern Recognit. Workshops (CVPRW)}, 2024, pp.~6536--6546.

\bibitem{ref36}
X. Wang, J. Li, J. Li, S. Wang, L. Yan, and Y. Xu, ``A collaborative network of Mamba and CNN for lightweight image super-resolution,''
\textit{IEEE Trans. Consum. Electron.}, vol.~71, no.~2, pp.~3591--3604, 2025, doi:~10.1109/TCE.2025.3572477.

\bibitem{ref44}
X. Di, L. Peng, P. Xia, \textit{et al.}, ``QMambaBSR: Burst image super-resolution with query state space model,''
in \textit{Proc. IEEE/CVF Conf. Comput. Vis. Pattern Recognit. (CVPR)}, 2025, pp.~23080--23090.

\bibitem{ref45}
Z. Xiao and X. Wang, ``Event-based video super-resolution via state space models,''
in \textit{Proc. IEEE/CVF Conf. Comput. Vis. Pattern Recognit. (CVPR)}, 2025, pp.~12564--12574.

\bibitem{ref28}
T. Dao and A. Gu, ``Transformers are SSMs: Generalized models and efficient algorithms through structured state space duality,''
in \textit{Proc. 41st Int. Conf. Mach. Learn. (ICML)}, 2024, pp.~10041--10071.

\bibitem{ref29}
P. Charbonnier, L. Blanc-F{\'e}raud, G. Aubert, \textit{et al.}, ``Deterministic edge-preserving regularization in computed imaging,''
\textit{IEEE Trans. Image Process.}, vol.~6, no.~2, pp.~298--311, 1997.

\bibitem{EDVR}
X. Wang, K.~C.~K. Chan, K. Yu, C. Dong, and C.~C. Loy, ``EDVR: Video restoration with enhanced deformable convolutional networks,''
in \textit{Proc. IEEE/CVF Conf. Comput. Vis. Pattern Recognit. Workshops (CVPRW)}, 2019, pp.~1954--1963, doi:~10.1109/CVPRW.2019.00247.

\bibitem{BasicVSRPP}
K.~C.~K. Chan, S. Zhou, X. Xu, and C.~C. Loy, ``BasicVSR++: Improving video super-resolution with enhanced propagation and alignment,''
in \textit{Proc. IEEE/CVF Conf. Comput. Vis. Pattern Recognit. (CVPR)}, 2022, pp.~5972--5981.

\bibitem{DBSR}
G. Bhat, M. Danelljan, L.~Van~Gool, and R. Timofte, ``Deep burst super-resolution,''
in \textit{Proc. IEEE/CVF Conf. Comput. Vis. Pattern Recognit. (CVPR)}, 2021, pp.~9209--9218.

\bibitem{MADNet}
Y. Xiao, Q. Yuan, K. Jiang, Y. Chen, S. Wang, and C.-W. Lin, ``Multi-axis feature diversity enhancement for remote sensing video super-resolution,''
\textit{IEEE Trans. Image Process.}, vol.~34, pp.~1766--1778, 2025, doi:~10.1109/TIP.2025.3547298.

\end{thebibliography}
\end{document}